\documentclass{article} 
\usepackage{iclr2027_conference,times}

\usepackage{hyperref}
\usepackage{url}
\usepackage[utf8]{inputenc}
\usepackage{microtype}
\usepackage{xspace}
\usepackage{xcolor}

\usepackage{amsmath,amssymb,amsfonts,bm}

\usepackage{graphicx}
\usepackage[font=small]{caption}
\usepackage{wrapfig}

\usepackage{booktabs}
\usepackage{multirow}

\usepackage{algorithm}
\usepackage{algpseudocode}

\hypersetup{
    breaklinks=true,
    colorlinks=true,
    citecolor=teal,
    linkcolor=purple,
    urlcolor=purple
}

\usepackage{color}
\usepackage{pifont}  

\providecommand{\omval}[2]{$#1^{\scriptscriptstyle\pm #2}$}
\providecommand{\ombest}[2]{$\mathbf{#1}^{\scriptscriptstyle\pm #2}$}
\providecommand{\omsecond}[2]{$\underline{#1}^{\scriptscriptstyle\pm #2}$}

\title{Harnessing Coupled Stream Completion \\ for Human–Object Interaction Modeling}

\author{
Dawei Guan\textsuperscript{1} \quad
Di Yang\textsuperscript{1} \quad
Jiangtao Wang\textsuperscript{1} \\
\textsuperscript{1} School of Artificial Intelligence and Data Science, University of Science and Technology of China \\
\href{https://visdyn.github.io/TRACE/}{\texttt{https://visdyn.github.io/TRACE/}}
}

\iclrfinalcopy

\begin{document}

\maketitle

\lhead{Preprint.}

\vspace{-0.7cm}

\begin{center}
    \centering
    \includegraphics[width=\textwidth]{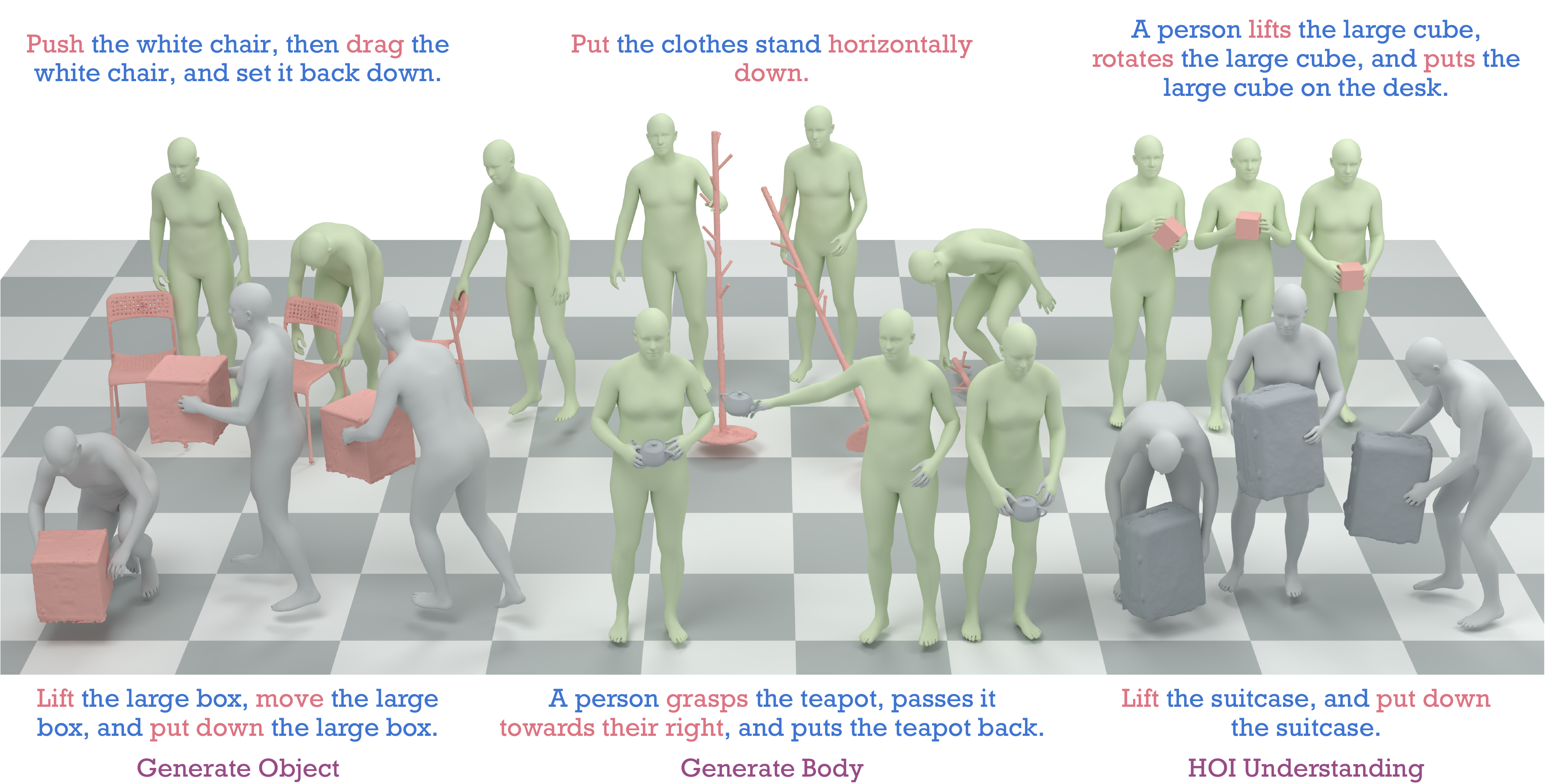}
    \captionof{figure}{
    Given a text prompt, our TRACE generates coherent human-object interaction
    motions via a coupled latent flow, while its structured representation
    further supports stream completion and HOI understanding.
    }
    \label{fig:teaser}
    \vspace{2em}
\end{center}

\begin{abstract}

Text-conditioned human-object interaction (HOI) generation requires body motion, object trajectories \& rotations, and hand articulation to remain coordinated. These components differ in scale and dynamics, but must agree on contact, relative pose, and timing. A shared representation may limit the distinct structure of each stream, while independent generation prevents each stream from responding to changes in the others. Latent supervision alone also does not directly constrain contact after decoding. We propose TRACE, a continuous latent framework that keeps stream states separate and couples their updates. TRACE encodes body, object, and hand motion into separate latents and predicts each stream velocity from the complete current interaction state. Geometric losses on decoded motion further constrain contact and object-relative motion over time. The same model supports completion of any single absent stream from the other two. Frozen flow features also serve as input to a language model for HOI understanding. Experiments on InterAct, OMOMO, and BEHAVE show that joint completion training improves generation and that frozen flow features improve understanding over raw-motion encoding. On InterAct, TRACE achieves the highest contact precision, recall, and F1 among the compared methods.

\end{abstract}
\section{Introduction}

Text-conditioned Human-Object Interaction (HOI) Generation synthesizes a 3D motion sequence in which a human interacts with a dynamic object according to a language description. Realistic HOI motion supports animation, virtual reality, gaming, and robotics simulation. Recent generative models have extended text-conditioned motion synthesis from isolated human actions to interactions with moving objects \citep{hoidiff,CG_HOI,chois,THOR,HOIAnimator,light}. Unlike ordinary text-to-motion, the object is not merely a static condition: its generated trajectory changes the feasible human motion throughout the sequence. HOI comprises body motion, object trajectories, and fine-grained hand articulation, which differ substantially in scale, dimensionality, and dynamics. Generating coherent interaction therefore requires preserving these distinct structures while aligning contact locations, relative poses, and interaction timing.

These heterogeneity raise a central question: how can each stream retain its own structure while all streams stay coordinated?
Recent generators coordinate motion through shared denoisers, separate branches, or auxiliary contact and relation guidance \citep{hoidiff,CG_HOI,THOR,HOIAnimator,rog,light}. 
Two design choices matter here: how to represent each stream and how to exchange information across streams. A shared representation allows information exchange but may limit the distinct structure of each stream. Separate representations preserve this structure, but each update still needs access to the current states of the other streams. With independent dynamics, each update depends only on its own stream and the shared conditions.
%
%
Even under the same text and geometry conditions, a body update cannot respond to the current object and hand states, and vice versa. The streams may therefore appear plausible individually while producing missed contact, object drift, or inconsistent interaction timing when combined.
Compressed latent spaces reduce temporal redundancy and provide a smoother generation domain \citep{mld,ardhoi,flowcomotion}, but compression alone does not determine how contact information should pass among streams. Latent regression also leaves contact, relations in the object frame, and interaction timing unconstrained after decoding. More importantly, stream decomposition exposes natural conditional relations: any one component can be inferred from others. Conditional HOI methods show that one motion stream can be inferred from others~\citep{omomo,DiffGrasp}. However, a  multi-stream model supporting both complete HOI generation and completion of any missing stream, within a shared temporal model remains underexplored.
Separate latent states need not imply independent generative dynamics. The distinction between \textit{state separation} and \textit{dynamic coupling} is essential. Body, object, and hand streams can retain specialized states and outputs, while each velocity depends on the complete current HOI state. Therefore, the model is joint in the information available to each update, but separate in what it represents and produces. When trajectories drift, every stream can respond to the current mismatch rather than only to static text and geometry conditions. Coordination is learned throughout generation instead of being imposed after motion is produced. Decomposing HOI into three streams naturally defines three completion tasks, each recovering one missing stream from the other two. These conditional relations define three single-missing-stream completion tasks within the same model. Features capable of predicting any missing stream are also encouraged to encode the action, manipulated object, and temporal progression, providing a natural semantic interface for understanding. 

We introduce \textbf{TRACE}, short for \textbf{T}ri-stream \textbf{R}epresentation \textbf{A}nd Stream \textbf{C}ompletion for Human-Object Interaction Mod\textbf{E}ling, a coupled continuous latent framework for text-conditioned Human-Object Interaction Generation, stream completion, and HOI understanding. As illustrated in Figure \ref{fig:teaser}, TRACE generates diverse human-object interactions from text and object geometry while supporting stream completion and HOI understanding. TRACE follows a separation-and-coupling principle. A split temporal VAE \citep{vae} maps body, hand, and object motion into separate but temporally aligned latents, preserving their distinct structures while reducing sequence length. These streams are not generated independently: a coupled latent flow \citep{flow_matching,rectified_flow} predicts every stream-specific velocity from the complete evolving HOI state, allowing their updates to respond to one another throughout generation. To ground latent transport in interaction geometry, clean latent estimates are decoded during training and supervised for contact, object-frame anchoring, and interaction timing. The same representation and flow backbone support Text-to-HOI generation and completion of any missing stream, while frozen coupled-flow states provide continuous memory tokens to a language model for HOI understanding.

    
    

Our contributions can be summarized into three parts:
\begin{itemize}
    \item We introduce TRACE, a novel coupled continuous latent framework for text-conditioned Human-Object Interaction Generation. 
    The coupled flow preserves separate body, object, and hand states, while each velocity depends on the complete current HOI state. Decoded geometric losses further constrain contact and object-relative motion over time.

    \item We reuse the coupled representation for completion and understanding. Completion of each missing stream is jointly trained with Text-to-HOI, while frozen flow features are bridged to a language model for HOI understanding.
    
    \item Experiments on three datasets evaluate generation, stream completion, and HOI understanding. Ablations quantify the benefits of stream coupling and decoded geometric supervision for contact consistency, along with the effect of the latent representation.

\end{itemize}
\section{Related Work}

\noindent\textbf{Text-driven HOI Generation.}
Text-driven HOI generation has progressed from hand-object synthesis to dynamic whole-body interaction. Text2HOI generates detailed hand-object motion, while InterDiff predicts future human-object interactions from motion history \citep{Text2HOI,interdiff}. Recent diffusion-based methods extend language control to body motion and dynamic objects through initial-state and waypoint conditioning, physics-informed correction, contact or affordance guidance, and relation-aware generation \citep{chois,hoidiff,CG_HOI,THOR,HOIAnimator}. Other methods broaden the task toward zero-shot synthesis, long-horizon instruction following with physics tracking, and generation beyond the training domain \citep{InterDreamer,CRFnet,ood_hoi}. More recent work further targets geometric fidelity, long-sequence consistency, and modality-aware generation \citep{rog,chainhoi,ardhoi,light}, reflecting growing attention to both interaction realism and controllability under increasingly complex human-object dynamics. These advances increasingly emphasize accurate coordination among articulated human motion, object trajectories, and contact events rather than treating object motion as a secondary condition. Despite this progress, synthesizing text-aligned body, object, and hand motion that remains spatially and temporally coherent throughout the interaction remains challenging.

\noindent\textbf{Motion Representation Learning.}
Motion generation has explored representations for preserving temporal structure and dependencies among motion components. Full-sequence representations jointly encode synchronized human-object trajectories \citep{interdiff,chois}, whereas factorized representations separate body, object, hand, or contact components \citep{Text2HOI,hoidiff,HOIAnimator,light}. Interaction representations introduce explicit relations through contact variables, geometric fields, or kinematic chains \citep{CG_HOI,THOR,rog,chainhoi}. Latent representations compress motion into compact generation spaces. MLD uses continuous VAE latents \citep{mld}, ARDHOI predicts continuous HOI tokens \citep{ardhoi}, and FlowCoMotion combines discrete tokens with continuous latents for text-conditioned flow \citep{flowcomotion}. Conditional modeling treats observed motion as input. OMOMO and DiffGrasp condition human motion on object trajectories, showing that partial observations constrain missing interaction dynamics \citep{omomo,DiffGrasp}. These formulations balance compactness, component specialization, and motion coordination. TriDi \citep{tridi} models static human-object distributions with one diffusion transformer, while Uni-HOI \citep{unihoi} and HOIGPT \citep{HOIGPT} use discrete representations for joint or bidirectional generation. Unlike LIGHT's asynchronous denoising, TRACE couples temporal body, object, and hand streams through continuous latent flow, with each velocity depending on the complete current state.

\noindent\textbf{Motion-Language Understanding.}
Motion-language understanding studies mappings between motion and text, including retrieval, understanding, and generation. Token-based methods discretize motion into language-like units for reciprocal translation and instruction following \citep{tm2t,motiongpt}. Alignment-based methods learn shared motion-text spaces \citep{lamp}, while MotionGPT3 \citep{motiongpt3} connects continuous motion latents to language parameters for bidirectional generation. Interaction-centered models extend this interface to human-scene and human-object interactions. HSI-GPT and HSI-GPT2 jointly model scene, motion, and language \citep{hsigpt,hsigpt2}, while Uni-HOI models a joint distribution over text, human, and object motion \citep{unihoi}. HOIGPT \citep{HOIGPT} represents hand-object sequences with discrete codes for bidirectional generation. These studies suggest that richer motion representations can provide language models with more explicit interaction semantics. In particular, representations learned from generation may encode contact evolution, object-relative motion, and coordinated body-hand dynamics that are directly useful for semantic description. Whether frozen human-object generation features retain reusable interaction semantics therefore remains underexplored.

\section{Methodology}
\label{sec:method}

\begin{figure*}
    \centering
    \includegraphics[width=\linewidth]{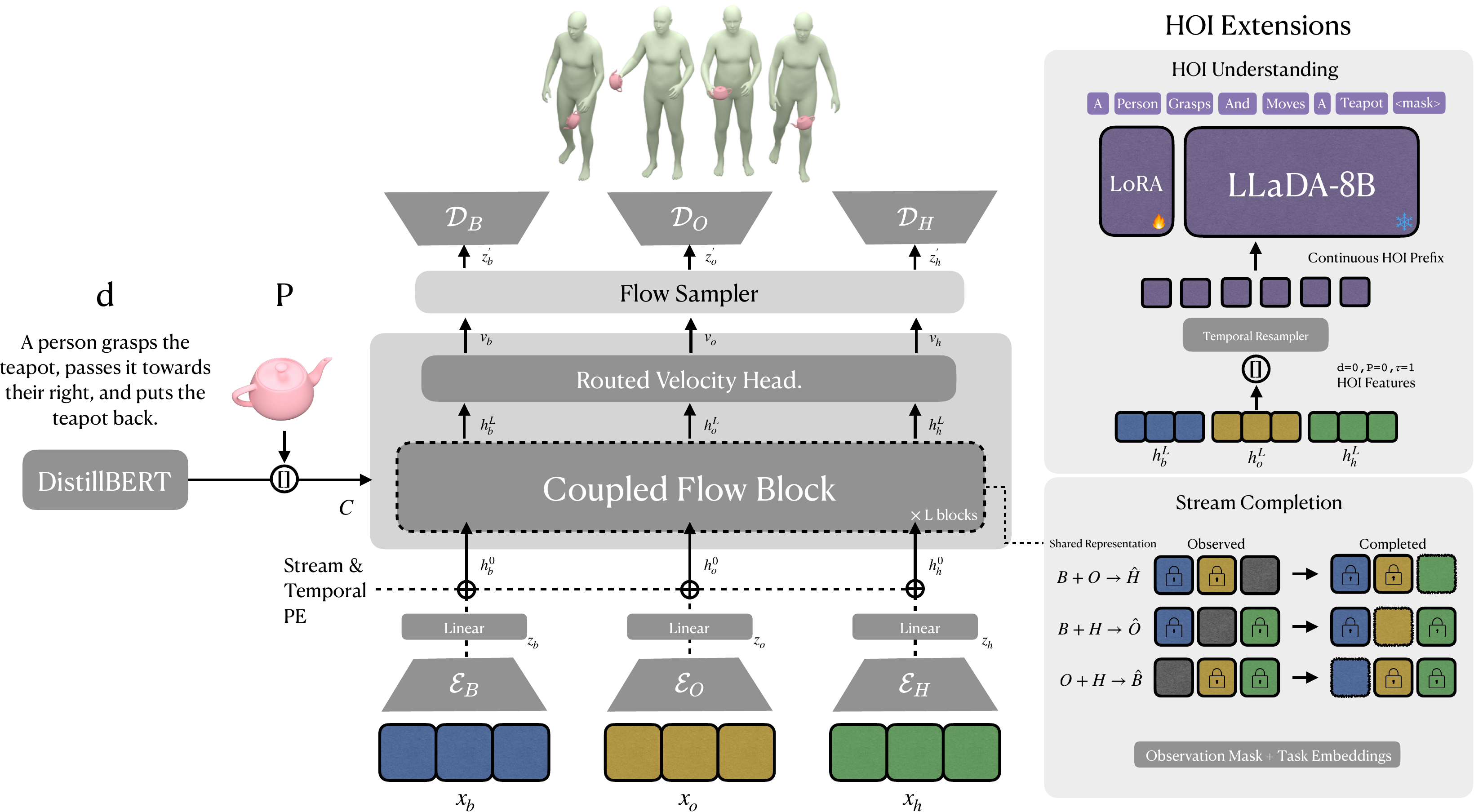}
    \caption{\footnotesize \textbf{Overview of TRACE.} \textit{Left: HOI Generation.} We form different modalities, e.g., body, hand, and object, each compressed into latent vectors by its own encoder. After adding stream and temporal embeddings, a coupled flow transformer predicts their velocities, which are integrated by a flow sampler. Then, different modalities' decoders reconstruct them into clean motion. \textit{Right: HOI Extensions.} Leveraging its HOI representation, we enable HOI understanding via language model fine-tuning. Joint training with three completion paths, each targeting one stream, adds completion capability while improving HOI Generation.}
    \label{fig:method}
\end{figure*}

\noindent \textbf{Overview.}
We formulate HOI generation as synthesizing a $T$-frame 3D motion sequence in which a human interacts with an object. As shown in Figure \ref{fig:method}, the input consists of a text description $\boldsymbol{d}$ and canonical object geometry $\boldsymbol{p}$ \citep{bps}, and the output is a synchronized sequence $\boldsymbol{x}=(\boldsymbol{x}_{b},\boldsymbol{x}_{h},\boldsymbol{x}_{o})$ of body, hand, and object streams.
Following SMPL-H \citep{smpl,MANO}, $\boldsymbol{x}_{b}\in\mathbb{R}^{T\times22\times3}$ contains 22 body joint positions, while $\boldsymbol{x}_{h}\in\mathbb{R}^{T\times120}$ contains 30 hand joint positions and rotations. The object stream $\boldsymbol{x}_{o}\in\mathbb{R}^{T\times9}$ contains a 6D rotation \citep{6D_rotation} and 3D translation.
TRACE encodes the three streams into separate continuous latents, jointly evolves them through a coupled flow, and decodes them into motion. It also completes any missing stream $\boldsymbol \hat{x}_{m}$, $m\in\{b,o,h\}$, from $\boldsymbol{d}$, $\boldsymbol{p}$, and the observed streams $\boldsymbol{x}_{-m}$. For HOI understanding, frozen flow hidden states are used to fine-tune a language model for interaction modeling.

\noindent\textbf{HOI Representation.}
Following \citet{Text2HOI}, we retain body, hand, and object motion as three explicit streams with distinct properties: articulated body motion, rigid object trajectories, and fine-grained hand movements. Separate encoders avoid forcing heterogeneous streams into one feature distribution, while temporal alignment preserves the frame-wise correspondence needed to recover contact and relative motion after decoding. The three streams share the same temporally aligned latent grid. The temporal VAE reduces its length by a factor \(s\). We therefore construct a split temporal variational autoencoder (VAE) \citep{vae}, with a separate encoder and decoder $(\mathcal E_m,\mathcal D_m)$ for each stream $m\in\{b,o,h\}$. The encoder compresses the original motion $\boldsymbol{x}_m\in\mathbb{R}^{t\times f_{m}}$ into a latent vector $\boldsymbol{z}_m\in\mathbb{R}^{l\times c_{m}}$, where $f_m$ and $c_m$ denote the input and latent dimensions of the stream, respectively, and $l=\lceil t/s\rceil$ with temporal downsampling factor $s$. The latent encoding and reconstruction are defined as:
\begin{equation}
q_m(\boldsymbol{z}_m\mid\boldsymbol{x}_m)
=
\mathcal{N}\!\left(
\boldsymbol{\mu}_m(\boldsymbol{x}_m),
\boldsymbol{\Sigma}_m(\boldsymbol{x}_m)
\right),
\qquad
\widehat{\boldsymbol{x}}_m
=
\mathcal D_m(\boldsymbol{z}_m).
\end{equation}
Here, $\boldsymbol{\mu}_m(\boldsymbol{x}_m)$ and $\boldsymbol{\Sigma}_m(\boldsymbol{x}_m)$ denote the mean and diagonal covariance predicted by $\mathcal E_m$, and $\widehat{\boldsymbol{x}}_m$ denotes the reconstructed motion produced by $\mathcal D_m$. 
To preserve the motion characteristics of each stream after temporal compression, we train the split VAE with the following objective: $\mathcal{L}_{\mathrm{vae}}=\mathcal{L}_{\mathrm{rec}}+\mathcal{L}_{\mathrm{reg}}$.
$\mathcal{L}_{\mathrm{rec}}$ denotes masked reconstruction over valid frames. The regularization term $\mathcal{L}_{\mathrm{reg}}$ aims to promote reasonable Human-Object motion reconstruction, and it consists of four components. Specifically, $\mathcal{L}_{\mathrm{reg}}$ combines object pose reconstruction, temporal velocity consistency, body kinematic consistency, and weak KL regularization, complementing masked reconstruction without overwhelming the continuous latent space. Full details are provided in Appendix \ref{sec:vae_objectives}.
After training, we freeze the VAE and the latent space produced by $\mathcal E_m$. Each mean is standardized independently for each channel as $\widetilde{\boldsymbol{z}}_m=(\boldsymbol{\mu}_m(\boldsymbol{x}_m)-\overline{\boldsymbol{\mu}}_m)/\overline{\boldsymbol{\sigma}}_m$, where $\overline{\boldsymbol{\mu}}_m$ and $\overline{\boldsymbol{\sigma}}_m$ denote the channel normalization parameters of the latent space. The resulting $\widetilde{\boldsymbol{z}}_m$ is then used as the target latent vector for flow matching modeling.

\noindent\textbf{Latent Flow Modeling.}
Given the latent vectors $\widetilde{\boldsymbol{z}}_m$ produced by the frozen VAE, we learn a direct flow from independent Gaussian sources to the conditional distribution of HOI latents of the three streams, following flow matching \citep{flow_matching,rectified_flow}. For each stream $m\in\{b,o,h\}$, we independently sample $\boldsymbol{z}_{m,0}\sim\mathcal{N}(\boldsymbol{0},\boldsymbol{I})$, while sharing the flow time $\tau\sim\mathcal{U}(0,1)$ across the three streams. The intermediate state and its target velocity along a linear path are defined as:
\begin{equation}
\boldsymbol{z}_{m,\tau}
=
(1-\tau)\boldsymbol{z}_{m,0}
+
\tau\widetilde{\boldsymbol{z}}_m,
\qquad
\boldsymbol{u}_{m,\tau}
=
\widetilde{\boldsymbol{z}}_m-\boldsymbol{z}_{m,0}.
\end{equation}
Let $\boldsymbol{z}_{\tau}=(\boldsymbol{z}_{b,\tau},\boldsymbol{z}_{o,\tau},\boldsymbol{z}_{h,\tau})$ denote the joint state; conditioned on $(\boldsymbol{z}_{\tau},\tau,\boldsymbol{d},\boldsymbol{p})$, the model predicts one velocity for each stream and trains them with:
\begin{equation}
\mathcal{L}_{\mathrm{fm}}
=
\sum_{m\in\{b,o,h\}}
\frac{
\left\|
\boldsymbol{M}\odot
\left(
\boldsymbol{v}_{\theta,m}
-
\boldsymbol{u}_{m,\tau}
\right)
\right\|_2^2
}{
c_m\left\|\boldsymbol{M}\right\|_1
}.
\end{equation}
Here, $\boldsymbol{M}$ is the temporal validity mask, broadcast across latent channels, and the denominator normalizes each stream by its valid latent elements.

Flow matching supervises latent velocity but not decoded interaction geometry. At time $\tau$, the predicted velocity yields a clean latent estimate
$\widehat{\boldsymbol{z}}_{m,1}=\boldsymbol{z}_{m,\tau}+(1-\tau)\boldsymbol{v}_{\theta,m}$.
The frozen VAE decodes the three latents while propagating gradients from decoded motion to the flow model. We further impose two interaction objectives: $\mathcal{L}_{\mathrm{c}}$ supervises human-object contact with BCE and Dice losses \citep{Text2HOI,InteractVLM}, while $\mathcal{L}_{\mathrm{A/E}}$ preserves object anchoring and temporal consistency. The contact term supervises whether interaction occurs, whereas the anchoring and episode terms constrain where it occurs in the object coordinate frame and how the relation evolves over time. The final objective is:
\begin{equation}
\mathcal{L}_{\mathrm{flow}}
=
\mathcal{L}_{\mathrm{fm}}
+
\lambda_{\mathrm{c}}\mathcal{L}_{\mathrm{c}}
+
\mathcal{L}_{\mathrm{A/E}}.
\end{equation}
It jointly supervises latent transport and decoded interaction geometry. Full details are provided in Appendix \ref{sec:training_supp}. These constraints improve motion quality and interaction consistency (Figure \ref{fig:base} and Table \ref{tab:interact_main}).

\noindent\textbf{Coupled Flow.}
Separating latent states does not require factorized transport dynamics. For $m\in\{b,o,h\}$, an independent stream flow evolves each latent only from its own current state:
\begin{equation}
\frac{\mathrm d\boldsymbol{z}_{m,\tau}}{\mathrm d\tau}=\boldsymbol{v}_{\theta,m}
(\boldsymbol{z}_{m,\tau},\tau,\boldsymbol{d},\boldsymbol{p}), \qquad
\frac{\partial\boldsymbol{v}_{\theta,m}}{\partial\boldsymbol{z}_{n,\tau}}=\boldsymbol{0}, \qquad 
n\neq m.
\label{eq5}
\end{equation}
Although shared textual and geometric conditions specify the interaction, they do not let one stream's velocity depend on other current states. Under independent Gaussian initialization, this induces conditionally factorized dynamics without a mechanism to reconcile body, object, and hand trajectories during transport. Thus, an object trajectory drifting from the hands cannot alter the body or hand updates, even though all streams share the same text and geometry conditions. We therefore couple the dynamics while preserving separate latent representations. 
TRACE evolves the separate stream latents through a single flow over the latent space $\mathcal{Z}=\mathcal{Z}_b\times\mathcal{Z}_o\times\mathcal{Z}_h$. Let $\boldsymbol{z}_{\tau}=(\boldsymbol{z}_{b,\tau},\boldsymbol{z}_{o,\tau},\boldsymbol{z}_{h,\tau})$ denote the complete HOI state. The coupled dynamics and responses between streams are:
\begin{equation}
\begin{aligned}
\frac{\mathrm d\boldsymbol{z}_{m,\tau}}{\mathrm d\tau}
=
\boldsymbol{v}_{\theta,m}
(\boldsymbol{z}_{\tau},\tau,\boldsymbol{d},\boldsymbol{p}), \qquad
\boldsymbol{J}_{\theta,mn}(\tau)
=
\frac{\partial\boldsymbol{v}_{\theta,m}}
{\partial\boldsymbol{z}_{n,\tau}},
\qquad m,n\in\{b,o,h\}.
\end{aligned}
\end{equation}
Here, $\boldsymbol{J}_{\theta,mn}$ measures how stream $m$ responds to stream $n$. Off-diagonal blocks need not vanish, enabling dependencies between streams. The vector field is joint in input but partitioned in output: each modality retains separate states and velocities while trajectories evolve jointly during transport. This joint input and separated output design couples transport without discarding the latent coordinates of each stream, allowing every frozen decoder to retain the motion statistics learned during VAE pretraining. Empirically, coupled flow improves motion quality and interaction consistency (Figure \ref{fig:streaming} and Table \ref{tab:ablation_stream_interact}).

\noindent\textbf{Architecture.}
The left of Figure~\ref{fig:method} illustrates the coupled flow transformer parameterizing $\boldsymbol{v}_{\theta}$.
(i) \emph{Latent vectors.} Each modal latent is linearly projected into a shared hidden dimension, then augmented with a learned stream embedding and sinusoidal temporal encoding:
$\boldsymbol{h}_{m}^{0}=\phi_m(\boldsymbol{z}_{m,\tau})+\boldsymbol{e}_{m}^{\mathrm{s}}+\boldsymbol{e}^{\mathrm{t}}$, $m\in\{b,o,h\}$.
(ii) \emph{Flow conditioning.} The shared flow time $\tau$ is sinusoidally embedded and combined with the text and geometry condition to modulate attention and feed-forward layers via AdaLN \citep{DiT}.
(iii) \emph{Text and geometry conditions.} The description $\boldsymbol{d}$ is encoded by frozen DistilBERT \citep{DistilBERT}, while object geometry $\boldsymbol{p}$ is represented by BPS \citep{bps} and projected into a geometry token. They are concatenated and injected into each stream via cross-attention in every transformer block.
(iv) \emph{Joint attention.}
Joint attention realizes the responses between streams $\boldsymbol{J}_{\theta,mn}$ required by the coupled flow. In each block $r$, the body, object, and hand streams form queries, keys, and values through separate projections. These features are concatenated for joint attention, split into individual streams, and processed by separate output projections and feed-forward networks:
\begin{equation}
\boldsymbol{h}_{m}^{r+1}
=
\mathcal{B}_{m}^{r}
\left(
\boldsymbol{h}_{b}^{r},
\boldsymbol{h}_{o}^{r},
\boldsymbol{h}_{h}^{r};
\tau,\boldsymbol{d},\boldsymbol{p}
\right),
\qquad
m\in\{b,o,h\}.
\end{equation}
Here, $\mathcal{B}_{m}^{r}$ denotes the update of stream $m$. Since each state depends on all streams, joint attention establishes a differentiable path from $\boldsymbol{z}_{n,\tau}$ to $\boldsymbol{v}_{\theta,m}$, enabling nonzero off-diagonal responses $\boldsymbol{J}_{\theta,mn}$.
(v) \emph{Routed velocity head.}
The velocity head maps coupled states to body, object, and hand velocities using a top-$k$ mixture-of-experts \citep{moe}. We use a shared top-$k$ router to produce a common mixture of stream-specific output projections for each interaction. A shared router predicts weights from the joint state and flow condition, while each expert uses stream-specific projections:
\begin{equation}
\boldsymbol{v}_{\theta,m}=\sum_{e\in\mathcal{K}(\boldsymbol{\pi})}\pi_e\,\mathcal{O}_{e,m}(\boldsymbol{h}_m),\qquad
\mathcal{L}_{\mathrm{r}}=E\sum_{e=1}^{E}\left(\overline{\pi}_e-\frac{1}{E}\right)^2.
\end{equation}
Here, $\mathcal{K}(\boldsymbol{\pi})$ denotes the selected experts, $\pi_e$ the normalized routing weight, and $\overline{\pi}_e$ its batch average. Shared routing weights mix all streams, while $\mathcal{O}_{e,m}$ remains stream-specific. The routing loss prevents expert concentration, giving
$\mathcal{L}=\mathcal{L}_{\mathrm{flow}}+\lambda_{\mathrm{r}}\mathcal{L}_{\mathrm{r}}$.

\noindent\textbf{Stream Completion.}
The coupled tri-stream representation allows TRACE to recover any missing stream. Unified conditional generators similarly share parameters across multimodal mappings \citep{OmniFlow,tridi}, while partial HOI sequences have been used to condition motion completion \citep{HOIGPT}. We jointly train standard HOI generation and three single-stream completion tasks. For a missing stream $m\in\{b,o,h\}$, $(\boldsymbol{x}_{-m},\boldsymbol{d},\boldsymbol{p})\mapsto\widehat{\boldsymbol{x}}_{m}$. All four tasks share the same VAE, coupled flow backbone, and sampling head. Joint training adds completion capability without requiring separate generators.

For each sample, observation and task embeddings identify the available streams and target $m$. Observed latents use $\tau_n=1$ for $n\neq m$, while the missing stream follows $\tau_m=\tau$. Joint attention infers the missing motion from the observed trajectories. To preserve the conditions, observed hidden states are restored after each block and their latents remain fixed throughout sampling. Joint training on all three completion tasks also improves HOI generation through stronger coupled flow coordination (Figure \ref{fig:stream_completion}, Table \ref{tab:interact_main}, \ref{tab:any_stream}, \ref{tab:omomo_main} and \ref{tab:behave_main}).

\noindent\textbf{HOI Understanding.}
For HOI understanding, we connect frozen coupled flow states to LLaDA and adapt it through LoRA \citep{lora} to describe a complete interaction. Given an interaction $\boldsymbol{x}$, the frozen VAE first maps its body, object, and hand streams into clean normalized latents. We read the frozen flow at $\tau=1$ under zero conditioning, which is included during flow training through $0.1$ condition dropout. The final hidden states before the velocity head form the HOI representation $\boldsymbol{h}=[\boldsymbol{h}_{b},\boldsymbol{h}_{o},\boldsymbol{h}_{h}]$. A learned readout embedding is added to each stream before the valid tokens are concatenated. A temporal slot resampler then compresses them into fixed memory tokens, followed by a projection into LLaDA's embedding space as a continuous prefix \citep{Llada,Lladav}:
\begin{equation}
\widehat{\boldsymbol{d}}
=
\mathcal{F}_{\mathrm{LM}}
\left(
\boldsymbol{q},
\mathcal{P}\!\left(\mathcal{R}(\boldsymbol{h})\right)
\right).
\end{equation}
Here, $\mathcal{R}$ and $\mathcal{P}$ denote the temporal resampler and prefix projection, while $\boldsymbol{q}$ contains the object category and learning query. The language branch reads frozen flow features without altering generation or completion. Full details on fine-tuning the language model are provided in Section \ref{sec:language_supp} of the Appendix. Empirically, harnessing the interaction relations learned during generation improves HOI understanding over directly encoding raw motion (Table \ref{tab:interact_understanding}).
\section{Experiments}

\noindent\textbf{Dataset.}
We mainly experiment on InterAct \citep{interact}, with ablations on its major subsets BEHAVE \citep{BEHAVE} and OMOMO \citep{omomo}. InterAct provides fine-grained textual annotations, and we follow its official train-test split. Since the data use both SMPL-H \citep{MANO} and SMPL-X \citep{smplx}, we convert all sequences to SMPL-H using the official conversion and consistently use SMPL-H joints.

\begin{figure*}
    \centering
    \includegraphics[width=\textwidth]{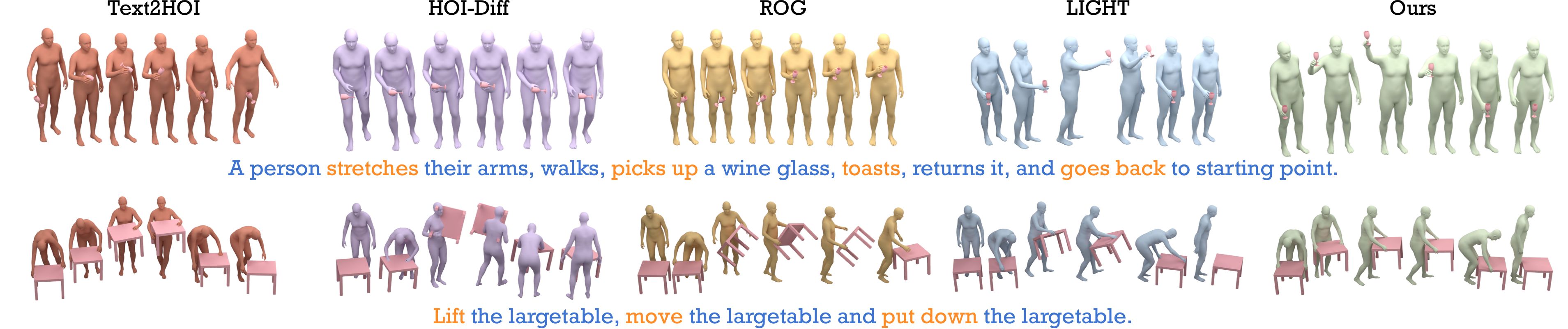}
    \caption{\textbf{Qualitative comparison} with baselines.  
    Our method yields more realistic human-object interactions, fewer contact/penetration artifacts, more accurate finger positioning, and better text-motion alignment.}
    \label{fig:base}
    \vspace{-1em}
\end{figure*}
\begin{table*}
\centering
\caption{\footnotesize
\textbf{Quantitative comparisons on the InterAct dataset between our method and baseline approaches.}
We report R-Precision with a batch size of 256.}
\label{tab:interact_main}
\resizebox{\textwidth}{!}{%
\begin{tabular}{lcccccccccccc}
\toprule
\multirow{2}{*}{Method}
& \multicolumn{3}{c}{R-Precision $\uparrow$}
& \multirow{2}{*}{FID $\downarrow$}
& \multirow{2}{*}{MM Dist. $\downarrow$}
& \multirow{2}{*}{Diversity $\rightarrow$}
& \multirow{2}{*}{FSR $\downarrow$}
& \multirow{2}{*}{Pene. $\downarrow$}
& \multirow{2}{*}{Contact $\rightarrow$}
& \multicolumn{3}{c}{Interaction $\uparrow$} \\
\cmidrule(lr){2-4}
\cmidrule(lr){11-13}
& Top 1 & Top 2 & Top 3
& & & & & &
& $C_{\mathrm{prec}}$
& $C_{\mathrm{rec}}$
& $C_{\mathrm{F1}}$ \\
\midrule

Ground Truth
& \omval{0.600}{0.004}
& \omval{0.834}{0.001}
& \omval{0.909}{0.001}
& \omval{0.000}{0.000}
& \omval{1.475}{0.003}
& \omval{7.781}{0.140}
& \omval{0.083}{0.204}
& \omval{0.076}{0.000}
& \omval{0.208}{0.000}
& \omval{1.000}{0.000}
& \omval{1.000}{0.000}
& \omval{1.000}{0.000} \\

\midrule

HOI-Diff
& \omval{0.413}{0.010}
& \omval{0.624}{0.009}
& \omval{0.740}{0.011}
& \omval{0.689}{0.031}
& \omval{3.029}{0.007}
& \omval{7.620}{0.085}
& \omval{0.072}{0.160}
& \ombest{0.103}{0.010}
& \omval{0.084}{0.006}
& \omval{0.722}{0.002}
& \omval{0.447}{0.026}
& \omval{0.501}{0.022} \\

CHOIS
& \omsecond{0.439}{0.005}
& \omsecond{0.660}{0.002}
& \omsecond{0.766}{0.003}
& \omval{0.572}{0.008}
& \omval{2.781}{0.017}
& \omsecond{7.717}{0.115}
& \omval{0.119}{0.234}
& \omval{0.131}{0.013}
& \omval{0.115}{0.002}
& \omval{0.710}{0.012}
& \omval{0.520}{0.004}
& \omval{0.541}{0.007} \\

InterDiff
& \ombest{0.501}{0.009}
& \ombest{0.722}{0.003}
& \ombest{0.824}{0.007}
& \omval{0.215}{0.001}
& \omval{2.461}{0.013}
& \omsecond{7.717}{0.115}
& \omval{0.092}{0.178}
& \omval{0.116}{0.011}
& \omval{0.124}{0.003}
& \omval{0.715}{0.000}
& \omval{0.562}{0.013}
& \omval{0.584}{0.003} \\

Text2HOI
& \omval{0.428}{0.003}
& \omval{0.637}{0.000}
& \omval{0.745}{0.000}
& \omval{0.331}{0.003}
& \omval{2.665}{0.002}
& \omval{7.631}{0.000}
& \omsecond{0.055}{0.123}
& \omsecond{0.105}{0.003}
& \omval{0.102}{0.001}
& \omval{0.711}{0.001}
& \omval{0.488}{0.000}
& \omval{0.532}{0.002} \\

ROG
& \omval{0.402}{0.007}
& \omval{0.613}{0.008}
& \omval{0.739}{0.009}
& \omval{0.326}{0.021}
& \omval{2.584}{0.018}
& \ombest{7.758}{0.076}
& \omval{0.058}{0.147}
& \omval{0.107}{0.005}
& \omval{0.168}{0.003}
& \omval{0.786}{0.005}
& \omval{0.646}{0.009}
& \omval{0.667}{0.007} \\

LIGHT
& \omval{0.421}{0.014}
& \omval{0.637}{0.016}
& \omval{0.754}{0.016}
& \ombest{0.148}{0.014}
& \omval{2.756}{0.018}
& \omval{7.712}{0.050}
& \omval{0.078}{0.197}
& \omval{0.132}{0.001}
& \omval{0.132}{0.001}
& \omval{0.731}{0.002}
& \omval{0.615}{0.016}
& \omval{0.627}{0.010} \\

\midrule

\text{TRACE}
& \omval{0.404}{0.003}
& \omval{0.615}{0.007}
& \omval{0.744}{0.005}
& \omval{0.199}{0.015}
& \ombest{2.256}{0.006}
& \omval{7.927}{0.051}
& \ombest{0.047}{0.146}
& \omval{0.117}{0.001}
& \omsecond{0.196}{0.001}
& \omsecond{0.830}{0.001}
& \omsecond{0.810}{0.002}
& \omsecond{0.799}{0.004} \\

\text{\text{TRACE}$^\text{ joint}$}
& \omval{0.407}{0.003}
& \omval{0.619}{0.008}
& \omval{0.755}{0.001}
& \omsecond{0.188}{0.007}
& \omsecond{2.262}{0.008}
& \omval{7.911}{0.063}
& \ombest{0.047}{0.148}
& \omval{0.111}{0.001}
& \ombest{0.198}{0.001}
& \ombest{0.835}{0.001}
& \ombest{0.811}{0.003}
& \ombest{0.806}{0.003} \\

\bottomrule
\end{tabular}%
}
\end{table*}

\noindent\textbf{Metrics.}
Following \citet{humanml3d}, we repeat evaluation 20 times and report mean ± 95\% confidence intervals. We evaluate HOI realism, diversity, text alignment, and physical plausibility. \textbf{FID} measures distribution similarity, \textbf{Diversity} measures motion variation, and \textbf{R-Precision} and \textbf{MM Dist} assess motion-text alignment. Following LIGHT \citep{light}, we report Foot Skating Ratio (\textbf{FSR}), Penetration Ratio (\textbf{Pene}), Contact Ratio (\textbf{Contact}), and frame-wise contact precision ($C_{prec}$), recall ($C_{rec}$), and F1 ($C_{F1}$). These complementary criteria capture distribution-level fidelity and interaction-level correctness, which is essential because visually plausible motions may still exhibit weak semantic alignment or physically implausible contact. We therefore report all metrics under the same evaluator and test split, enabling consistent comparisons across generation, completion, and understanding settings in our experiments. For HOI understanding, following \citet{motiongpt}, we use \textbf{Bleu@1/4}, \textbf{Rouge}, \textbf{Cider}, and \textbf{BertScore} to evaluate generated descriptions, together with \textbf{R-Precision} and \textbf{MM Dist} for motion-text alignment.

\noindent\textbf{Implementation Details.}
Object geometry is represented by a $1024$-point BPS \citep{bps}. The split VAE uses temporal downsampling $s=4$ and latent dimensions $(c_b,c_o,c_h)=(128,64,128)$. The coupled-flow backbone is a $12$ layer transformer with hidden dimension $1024$ and $16$ heads. The velocity head uses $4$ experts with top-$2$ routing. Models are trained on one NVIDIA A800 GPU with BF16. The language branch uses a two-layer temporal resampler with $32$ memory tokens and rank $8$ LoRA. Training uses AdamW with cosine learning-rate decay and gradient clipping for stable optimization of the coupled-flow backbone. We keep the VAE frozen during flow training and optimize the flow model jointly. For understanding, only the resampler, LoRA parameters, and output projection are updated, leaving the generator unchanged. Additional loss weights and schedules are provided in the Appendix \ref{sec:training_config}.

\noindent\textbf{Baselines.}
We compare against six recent HOI generation methods. HOI-Diff \citep{hoidiff} uses transformer-based diffusion with affordance-guided sampling. CHOIS \citep{chois} is adapted to text-only conditioning by removing state and waypoint inputs. InterDiff \citep{interdiff} replaces its motion-history encoder with a text encoder. Text2HOI \citep{Text2HOI} is extended to full-body HOI while retaining its static contact-map estimator. ROG \citep{rog} combines motion diffusion with relation guidance and is retrained on InterAct. LIGHT \citep{light} follows its original setting. All methods use the same preprocessing, split, and evaluator, while adapted baselines retain their original objectives and training schedules. For HOI understanding, we compare with HOIGPT \citep{HOIGPT}, TM2T \citep{tm2t}, MotionGPT \citep{motiongpt}, and LaMPM2T \citep{lamp}. HOIGPT is extended with a body stream, while other baselines only adapt their motion encoders to our datasets. All models are retrained on InterAct with their original understanding architectures and objectives. TRACE Raw replaces the frozen VAE-flow encoder with a frame projection of normalized motion, followed by the same temporal resampler and prefix projection. Both variants share object prompts, LLaDA/LoRA configurations, language objectives, and training schedules, isolating frozen coupled-flow features from raw-motion encoding.

\noindent\textbf{Quantitative Evaluation.}
As shown in Table \ref{tab:interact_main}, TRACE achieves the strongest interaction accuracy: joint training produces the highest contact precision, recall, and F1, while our variants yield the lowest MM Dist and FSR. Table \ref{tab:interact_understanding} shows that TRACE performs best on R@1, MM Dist, BLEU@4, and CIDEr for HOI understanding, with coupled-flow features outperforming raw-motion encoding on all metrics. Tables \ref{tab:omomo_main}, \ref{tab:behave_main}, \ref{tab:omomo_understanding}, and \ref{tab:behave_understanding} report generation and understanding on OMOMO and BEHAVE, confirming the benefits of joint training and flow-based HOI representations.

\noindent\textbf{Qualitative Evaluation.}
As shown in Figure~\ref{fig:base}, TRACE better follows descriptions while maintaining coordinated human-object interaction and stable contact, whereas baselines often omit actions or exhibit object drift, penetration, and floating artifacts. Figure \ref{fig:understanding} in the Appendix further shows the advantage of the complete tri-stream representation for HOI understanding.

\begin{figure*}
    \centering
    \includegraphics[width=\textwidth]{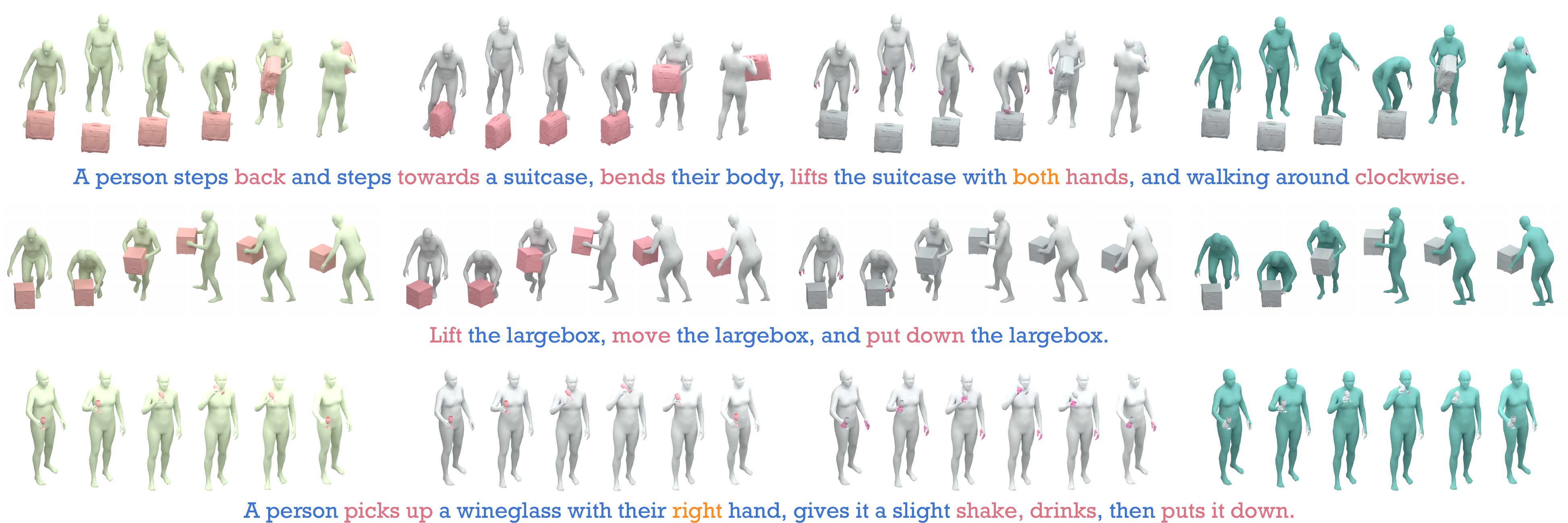}
    \caption{\footnotesize \textbf{Qualitative results of stream completion.} From left to right, we show the ground-truth interaction and the completion of object, hand, and body streams. Gray denotes observed motion, while pink, magenta, and teal denote the generated object, hands, and body, respectively.}
    \label{fig:stream_completion}
    \vspace{-1em}
\end{figure*}
\begin{table*}
    \caption{\footnotesize \textbf{Comparison of HOI understanding on InterAct.}
    Our model is trained for 100 epochs.}
    \label{tab:interact_understanding}
    \centering
    \resizebox{\textwidth}{!}{%
    \begin{tabular}{l c c c c c c c c c}
    \toprule
    \multirow[c]{2}{*}{Methods}
    & \multicolumn{3}{c}{R-Precision $\uparrow$}
    & \multirow[c]{2}{*}{MM Dist. $\downarrow$}
    & \multirow[c]{2}{*}{Bleu@1 $\uparrow$}
    & \multirow[c]{2}{*}{Bleu@4 $\uparrow$}
    & \multirow[c]{2}{*}{Rouge $\uparrow$}
    & \multirow[c]{2}{*}{Cider $\uparrow$}
    & \multirow[c]{2}{*}{BertScore $\uparrow$} \\
    \cmidrule(lr){2-4}
    & Top 1 & Top 2 & Top 3
    & & & & & & \\
    \midrule

    Real
    & 0.592 & 0.757 & 0.853
    & 1.486
    & --
    & --
    & --
    & --
    & -- \\

    \midrule

    TM2T
    & 0.206 & 0.351 & 0.466
    & 3.342
    & 35.18
    & 24.12
    & 56.17
    & 2.61
    & 52.16 \\

    MotionGPT
    & 0.224 & 0.391 & 0.504
    & 3.086
    & 39.86
    & 29.63
    & 60.84
    & 3.42
    & 57.83 \\

    LaMPM2T
    & 0.243 & 0.415 & 0.532
    & 2.894
    & 44.37
    & 34.42
    & 66.14
    & 4.09
    & 63.18 \\

    HOIGPT
    & \underline{0.256}
    & \textbf{0.452}
    & \textbf{0.570}
    & \underline{2.756}
    & \textbf{50.31}
    & \underline{39.22}
    & \textbf{71.08}
    & \underline{4.89}
    & \textbf{69.05} \\

    \midrule

    \text{\text{TRACE}$^\text{ Raw}$}
    & 0.232 
    & 0.408
    & 0.512 
    & 3.020
    & 41.57 
    & 30.82 
    & 62.24 
    & 3.55 
    & 58.81 \\

    \text{TRACE}
    & \textbf{0.260}
    & \underline{0.449}
    & \underline{0.567}
    & \textbf{2.742}
    & \underline{49.87}
    & \textbf{39.54}
    & \underline{70.18}
    & \textbf{4.95}
    & \underline{67.47} \\

    \bottomrule
    \end{tabular}%
    }
    \vspace{-2mm}
\end{table*}

\noindent\textbf{Stream Completion.}
As shown in Figure \ref{fig:stream_completion}, TRACE completes a missing body, object, or hand stream from the other two observed streams. For body completion, the generated posture follows the fixed hand-object trajectories and preserves their spatial constraints over time. Object completion recovers a rigid trajectory consistent with the observed human motion, while hand completion aligns fine-grained movements with the body and object. In all three settings, the missing component is inferred from cross-stream dependencies rather than generated independently. The completed component remains coordinated with the observed trajectories, while the observed motion is preserved. Table \ref{tab:any_stream} in the appendix reports stream-specific reconstruction and physical metrics on OMOMO, BEHAVE, and InterAct.

\begin{figure*}[t!]
    \centering
    \includegraphics[width=\textwidth]{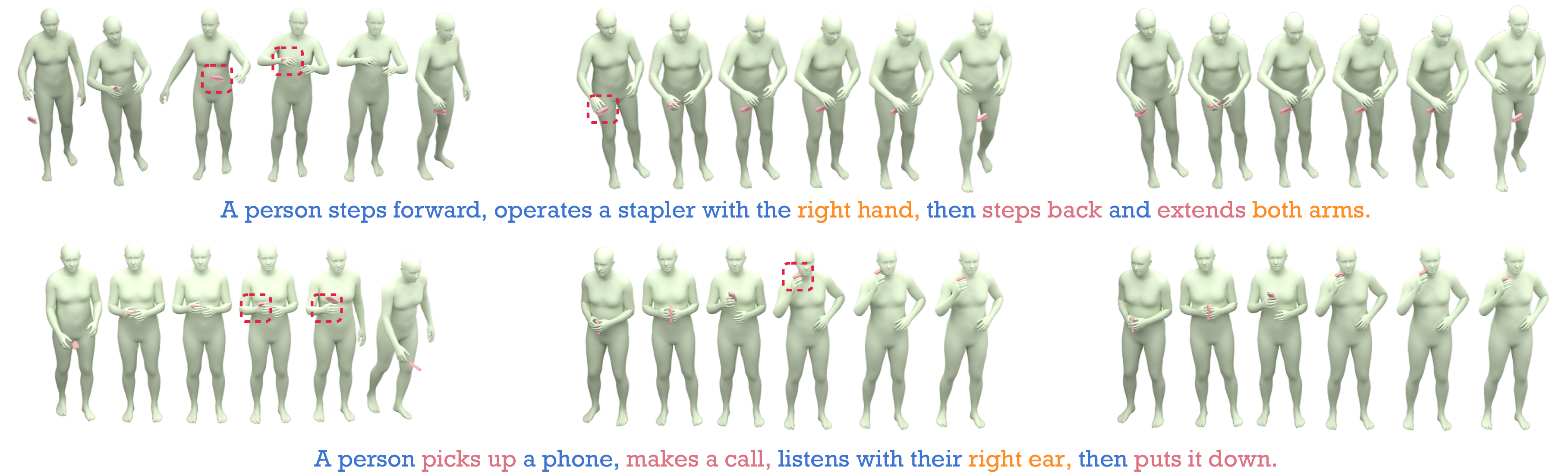}
    \caption{\textbf{Qualitative comparison of stream modeling.} We compare unified, independent, and coupled stream modeling. Red boxes highlight object drift and hand-object misalignment in the first two variants, while coupled modeling yields coherent interactions.}
    \label{fig:streaming}
    \vspace{-0.5em}
\end{figure*}

\noindent\textbf{Impact of Stream Modeling.}
We compare unified, independent, and coupled stream modeling. As shown in Table \ref{tab:ablation_stream_interact}, the coupled model performs best in R-Precision, FID, MM Dist, FSR, Contact, and all three interaction metrics. These results show that neither merging the streams nor evolving them independently captures their interaction dependencies as effectively as coupled flow. Figure \ref{fig:streaming} further shows that the alternative strategies often produce object drift or misaligned hand-object contact, whereas coupled modeling maintains coordinated trajectories throughout the interaction. Tables \ref{tab:ablation_stream_omomo} and \ref{tab:ablation_stream_behave} in the Appendix exhibit the same overall trend on OMOMO and BEHAVE.

\begin{table*}[h!]
    \caption{\footnotesize
    \textbf{Ablation study of stream modeling on InterAct.} We report R-Precision with a batch size of 256.}
    \label{tab:ablation_stream_interact}
    \centering
    \scriptsize
    \setlength{\tabcolsep}{2.6pt}
    \renewcommand{\arraystretch}{1.05}
    \resizebox{\textwidth}{!}{%
    \begin{tabular}{@{}l*{12}{c}@{}}
        \toprule
        \multirow[c]{2}{*}{Strategy}
        & \multicolumn{3}{c}{R-Precision $\uparrow$}
        & \multirow[c]{2}{*}{FID $\downarrow$}
        & \multirow[c]{2}{*}{MM Dist. $\downarrow$}
        & \multirow[c]{2}{*}{Diversity $\rightarrow$}
        & \multirow[c]{2}{*}{FSR $\downarrow$}
        & \multirow[c]{2}{*}{Pene. $\downarrow$}
        & \multirow[c]{2}{*}{Contact $\rightarrow$}
        & \multicolumn{3}{c}{Interaction $\uparrow$} \\
        \cmidrule(lr){2-4}
        \cmidrule(lr){11-13}
        & Top 1
        & Top 2
        & Top 3
        & & & & & & &
        $C_{\mathrm{prec}}$
        & $C_{\mathrm{rec}}$
        & $C_{F1}$ \\
        \midrule

        GT
        & \omval{0.600}{0.004}
        & \omval{0.834}{0.001}
        & \omval{0.909}{0.001}
        & \omval{0.000}{0.000}
        & \omval{1.475}{0.003}
        & \omval{7.781}{0.140}
        & \omval{0.083}{0.204}
        & \omval{0.076}{0.000}
        & \omval{0.208}{0.000}
        & \omval{1.000}{0.000}
        & \omval{1.000}{0.000}
        & \omval{1.000}{0.000} \\

        \midrule

        Independent
        & \omval{0.383}{0.005}
        & \omval{0.584}{0.007}
        & \omval{0.710}{0.005}
        & \omval{0.261}{0.017}
        & \omval{2.447}{0.008}
        & \omval{8.003}{0.056}
        & \omsecond{0.052}{0.109}
        & \ombest{0.103}{0.001}
        & \omval{0.146}{0.002}
        & \omval{0.756}{0.003}
        & \omval{0.579}{0.003}
        & \omval{0.606}{0.003} \\

        Unified
        & \omsecond{0.400}{0.004}
        & \omsecond{0.609}{0.006}
        & \omsecond{0.737}{0.004}
        & \omsecond{0.213}{0.016}
        & \omsecond{2.306}{0.007}
        & \omval{7.896}{0.054}
        & \omval{0.059}{0.133}
        & \omval{0.130}{0.001}
        & \omsecond{0.186}{0.002}
        & \omsecond{0.818}{0.002}
        & \omsecond{0.764}{0.002}
        & \omsecond{0.760}{0.002} \\

        Coupled
        & \ombest{0.404}{0.003}
        & \ombest{0.615}{0.007}
        & \ombest{0.744}{0.005}
        & \ombest{0.199}{0.015}
        & \ombest{2.256}{0.006}
        & \omval{7.927}{0.051}
        & \ombest{0.047}{0.146}
        & \omsecond{0.117}{0.001}
        & \ombest{0.196}{0.001}
        & \ombest{0.830}{0.001}
        & \ombest{0.810}{0.002}
        & \ombest{0.799}{0.004} \\

        \bottomrule
    \end{tabular}%
    }
\end{table*}
\begin{table*}[h!]
    \caption{\footnotesize
    \textbf{Ablation study of latent representation on InterAct.}
    We report R-Precision with a batch size of 256.}
    \label{tab:ablation_space}
    \centering
    \scriptsize
    \setlength{\tabcolsep}{2.6pt}
    \renewcommand{\arraystretch}{1.05}
    \resizebox{\textwidth}{!}{%
    \begin{tabular}{@{}l*{12}{c}@{}}
        \toprule
        \multirow[c]{2}{*}{Space}
        & \multicolumn{3}{c}{R-Precision $\uparrow$}
        & \multirow[c]{2}{*}{FID $\downarrow$}
        & \multirow[c]{2}{*}{MM Dist. $\downarrow$}
        & \multirow[c]{2}{*}{Diversity $\rightarrow$}
        & \multirow[c]{2}{*}{FSR $\downarrow$}
        & \multirow[c]{2}{*}{Pene. $\downarrow$}
        & \multirow[c]{2}{*}{Contact $\rightarrow$}
        & \multicolumn{3}{c}{Interaction $\uparrow$} \\
        \cmidrule(lr){2-4}
        \cmidrule(lr){11-13}
        & Top 1
        & Top 2
        & Top 3
        & & & & & & &
        $C_{\mathrm{prec}}$
        & $C_{\mathrm{rec}}$
        & $C_{F1}$ \\
        \midrule

        GT
        & \omval{0.600}{0.004}
        & \omval{0.834}{0.001}
        & \omval{0.909}{0.001}
        & \omval{0.000}{0.000}
        & \omval{1.475}{0.003}
        & \omval{7.781}{0.140}
        & \omval{0.083}{0.204}
        & \omval{0.076}{0.000}
        & \omval{0.208}{0.000}
        & \omval{1.000}{0.000}
        & \omval{1.000}{0.000}
        & \omval{1.000}{0.000} \\

        \midrule

        Raw & 
        \omsecond{0.399}{0.003} & \omsecond{0.607}{0.006} & \omsecond{0.735}{0.004} & \omsecond{0.228}{0.014} & \omsecond{2.314}{0.007} & \omval{7.918}{0.052} & \omsecond{0.055}{0.132} & \omsecond{0.135}{0.001} & \omsecond{0.182}{0.002} & \omsecond{0.809}{0.002} & \omsecond{0.750}{0.003} & \omsecond{0.748}{0.003} \\

        VQ 
        & \omval{0.384}{0.004}
        & \omval{0.586}{0.006}
        & \omval{0.711}{0.004}
        & \omval{0.269}{0.016}
        & \omval{2.406}{0.007}
        & \omval{7.884}{0.053}
        & \omval{0.063}{0.001}
        & \omval{0.146}{0.184}
        & \omval{0.158}{0.002}
        & \omval{0.771}{0.002}
        & \omval{0.681}{0.003}
        & \omval{0.685}{0.003} \\

        VAE & 
        \ombest{0.404}{0.003} & 
        \ombest{0.615}{0.007} & 
        \ombest{0.744}{0.005} & 
        \ombest{0.199}{0.015} & 
        \ombest{2.256}{0.006} & 
        \omval{7.927}{0.051} & 
        \ombest{0.047}{0.146} & 
        \ombest{0.117}{0.001} & 
        \ombest{0.196}{0.001} & 
        \ombest{0.830}{0.001} & 
        \ombest{0.810}{0.002} & 
        \ombest{0.799}{0.004} \\

        \bottomrule
    \end{tabular}%
    }
\end{table*}

\noindent\textbf{Impact of Latent Representation.}
We compare raw motion, discrete VQ latents, and continuous VAE latents under the same coupled-flow architecture. Table \ref{tab:ablation_space} shows that the VAE achieves the best R-Precision, FID, MM Dist, FSR, penetration, contact, and interaction metrics. Raw motion also outperforms VQ on most metrics, suggesting that discrete quantization discards fine-grained interaction cues. By preserving continuous stream-specific information while reducing temporal redundancy, the VAE provides a more effective space for coupled flow modeling.

\vspace{-0.5em}
\begin{wraptable}{r}{0.7\textwidth}
    \centering
    \caption{\footnotesize
    \textbf{Ablation study of geometric constraints on InterAct.} We report R-Precision with a batch size of 256.}
    \label{tab:geometric_constraints}
    \scriptsize
    \setlength{\tabcolsep}{2.4pt}
    \renewcommand{\arraystretch}{1.05}
    \resizebox{0.67\textwidth}{!}{%
    \begin{tabular}{@{}cc*{8}{c}@{}}
        \toprule
        \multirow[c]{2}{*}{$\mathcal{L}_{\mathrm{c}}$}
        & \multirow[c]{2}{*}{$\mathcal{L}_{\mathrm{A/E}}$}
        & \multirow[c]{2}{*}{R@3 $\uparrow$}
        & \multirow[c]{2}{*}{FID $\downarrow$}
        & \multirow[c]{2}{*}{FSR $\downarrow$}
        & \multirow[c]{2}{*}{Pene. $\downarrow$}
        & \multirow[c]{2}{*}{Contact $\rightarrow$}
        & \multicolumn{3}{c}{Interaction $\uparrow$} \\
        \cmidrule(lr){8-10}
        & & & & & & &
        $C_{\mathrm{prec}}$
        & $C_{\mathrm{rec}}$
        & $C_{F1}$ \\
        \midrule

        $\times$ & $\times$
        & \omval{0.736}{0.004}
        & \omval{0.220}{0.016}
        & \omval{0.053}{0.138}
        & \omval{0.137}{0.002}
        & \omval{0.173}{0.002}
        & \omval{0.796}{0.003}
        & \omval{0.708}{0.004}
        & \omval{0.715}{0.004} \\

        $\checkmark$ & $\times$
        & \omsecond{0.740}{0.003}
        & \omsecond{0.208}{0.015}
        & \omsecond{0.050}{0.104}
        & \omsecond{0.126}{0.001}
        & \omsecond{0.187}{0.002}
        & \omsecond{0.814}{0.002}
        & \omsecond{0.770}{0.003}
        & \omsecond{0.765}{0.003} \\

        $\checkmark$ & $\checkmark$
        & \ombest{0.744}{0.005}
        & \ombest{0.199}{0.015}
        & \ombest{0.047}{0.146}
        & \ombest{0.117}{0.001}
        & \ombest{0.196}{0.001}
        & \ombest{0.830}{0.001}
        & \ombest{0.810}{0.002}
        & \ombest{0.799}{0.004} \\

        \bottomrule
    \end{tabular}%
    }
    \vspace{-0.5em}
\end{wraptable}

\noindent\textbf{Impact of Geometric Constraints.}
Table \ref{tab:geometric_constraints} evaluates the geometric constraints applied to the decoded interaction. Adding $\mathcal{L}_{\mathrm{c}}$ to the flow objective consistently improves R@3, FID, FSR, penetration, and contact accuracy. Further introducing $\mathcal{L}_{\mathrm{A/E}}$ achieves the best result on every reported metric. This progression shows that contact supervision improves local alignment, while object-local anchoring and temporal consistency further stabilize interaction geometry over time.
\section{Conclusion}
We introduced TRACE, a structured continuous-latent framework for Text-to-HOI generation. TRACE preserves separate body, object, and hand states while conditioning each stream-specific velocity on the complete evolving HOI state; decoded-space geometric constraints further ground latent generation in contact locations, object-local relations, and interaction timing. The same generator handles Text-to-HOI and three stream completion tasks, while frozen interaction features support HOI understanding through a language bridge. Experiments across three datasets demonstrate generation and interaction performance, completion capability, and consistent understanding gains over raw-motion encoding. Together, these results show that structured cross-stream coupling provides a unified representation for coordinated generation, conditional reconstruction, and semantic interpretation across diverse human-object interaction tasks.

\bibliography{iclr2027_conference}
\bibliographystyle{iclr2027_conference}

\clearpage
\appendix

\setcounter{table}{0}
\renewcommand{\thetable}{\Alph{table}}
\renewcommand*{\theHtable}{\thetable}

\setcounter{figure}{0}
\renewcommand{\thefigure}{\Alph{figure}}
\renewcommand*{\theHfigure}{\thefigure}

\setcounter{section}{0}
\renewcommand{\thesection}{\Alph{section}}
\renewcommand*{\theHsection}{\thesection}

\section*{Appendix}

\noindent In this Appendix, we provide additional details of the training objectives, completion training, and language adaptation in Sec.~\ref{sec:method_supp}, followed by additional experimental results in Sec.~\ref{sec:add_exp_supp}.

\noindent \textbf{LLM Usage.} We employ large language models (LLMs), such as ChatGPT, to assist in polishing our paper. Specifically, LLMs are used to correct grammatical errors, refine word choice, and improve overall fluency. We do not use LLMs to formulate our methodology or run experiments.

\section{Additional Method Details}
\label{sec:method_supp}

This section first presents the objectives and training procedure of the generation and completion model, followed by details of the language adaptation branch.

\subsection{Generation and Completion Training}
\label{sec:training_supp}

Using the frozen VAE, we independently train the generation-only and joint generation-completion TRACE variants for 1000 epochs under the same optimization budget.

\paragraph{VAE Objectives.}
\phantomsection
\label{sec:vae_objectives}
The split VAE is trained using masked reconstruction together with four motion regularizers. All losses are averaged over valid frames. The reconstruction term combines the complete motion and object-stream errors, while the regularization term supervises object pose, temporal velocity, body kinematics, and latent distribution:
\begin{equation}
\begin{aligned}
\mathcal{L}_{\mathrm{rec}}
&=
\mathcal{L}_{1}(\widehat{\boldsymbol{x}},\boldsymbol{x})
+
\lambda_{\mathrm{o}}
\mathcal{L}_{1}(\widehat{\boldsymbol{x}}_{o},\boldsymbol{x}_{o}),\\
\mathcal{L}_{\mathrm{reg}}
&=
\lambda_{\mathrm{obj}}\mathcal{L}_{\mathrm{obj}}
+
\lambda_{\mathrm{vel}}\mathcal{L}_{\mathrm{vel}}
+
\lambda_{\mathrm{kin}}\mathcal{L}_{\mathrm{kin}}
+
\beta\mathcal{L}_{\mathrm{KL}}.
\end{aligned}
\end{equation}
Here, $\mathcal{L}_{\mathrm{obj}}$ applies Smooth L1 loss to object translation and a rotation loss $1-(\operatorname{tr}(\widehat{\boldsymbol{R}}^{\top}\boldsymbol{R})-1)/2$. $\mathcal{L}_{\mathrm{vel}}$ matches first-order differences of human joint positions and object translation, while $\mathcal{L}_{\mathrm{kin}}$ matches the body bone vectors defined by the SMPL-H kinematic tree. $\mathcal{L}_{\mathrm{KL}}$ is averaged over the valid latent positions of the three streams. We set $\lambda_{\mathrm{o}}=0.25$, $\lambda_{\mathrm{obj}}=0.25$, $\lambda_{\mathrm{vel}}=0.05$, $\lambda_{\mathrm{kin}}=0.1$, and $\beta=10^{-4}$.

\paragraph{Decoded Interaction Constraints.}
\phantomsection
\label{sec:decoded_constraints}
Given the clean latent estimate $\widehat{\boldsymbol{z}}_{m,1}$, the frozen VAE reconstructs the complete interaction $\widehat{\boldsymbol{x}}$. Let $d_{t,j}$ and $\widehat{d}_{t,j}$ denote the ground-truth and predicted distances from human joint $j$ to the object surface at frame $t$. We define the contact label and its differentiable prediction as $c_{t,j}=\mathbb{1}[d_{t,j}<\gamma]$ and $\widehat{c}_{t,j}=\operatorname{sigmoid}((\gamma-\widehat{d}_{t,j})/\sigma_{\mathrm{c}})$, respectively. The contact objective is
\begin{equation}
\mathcal{L}_{\mathrm{c}}
=
\mathcal{L}_{\mathrm{BCE}}(\widehat{\boldsymbol{c}},\boldsymbol{c})
+
2\mathcal{L}_{\mathrm{Dice}}(\widehat{\boldsymbol{c}},\boldsymbol{c}),
\end{equation}
where positive contacts receive a weight of $3$. We use $\gamma=0.05$\,m and $\sigma_{\mathrm{c}}=0.01$\,m.

To supervise relative interaction geometry, let $\boldsymbol{a}_{t,j}$ be a human anchor and define its coordinates in the object frame as $\boldsymbol{r}_{t,j}=\boldsymbol{R}_{t}^{\top}(\boldsymbol{a}_{t,j}-\boldsymbol{t}_{t})$. A soft contact weight $w_{t,j}$ selects anchors close to the ground-truth object. Using Smooth L1 loss $\rho(\cdot)$, the anchoring and temporal episode objectives are
\begin{equation}
\mathcal{L}_{\mathrm{A}}
=
\frac{\sum_{t,j}w_{t,j}\,
\rho\!\left((\widehat{\boldsymbol{r}}_{t,j}-\boldsymbol{r}_{t,j})/s_{\mathrm{A}}\right)}
{\sum_{t,j}w_{t,j}},
\end{equation}
\begin{equation}
\mathcal{L}_{\mathrm{E}}
=
\frac{1}{|\mathcal{S}|}
\sum_{s\in\mathcal{S}}
\frac{\sum_{t,j}w_{t,j}w_{t+s,j}\,
\rho\!\left((\Delta_s\widehat{\boldsymbol{r}}_{t,j}-\Delta_s\boldsymbol{r}_{t,j})/s_{\mathrm{E}}\right)}
{\sum_{t,j}w_{t,j}w_{t+s,j}},
\end{equation}
where $\Delta_s\boldsymbol{r}_{t,j}=\boldsymbol{r}_{t+s,j}-\boldsymbol{r}_{t,j}$ and $\mathcal{S}=\{1,4,8\}$. We use $\mathcal{L}_{\mathrm{A/E}}=0.01\mathcal{L}_{\mathrm{A}}+0.005\mathcal{L}_{\mathrm{E}}$, with $s_{\mathrm{A}}=0.05$\,m and $s_{\mathrm{E}}=0.02$\,m. These terms constrain both the contact location in the object frame and its evolution over time.

\paragraph{Joint Generation and Completion.}
We represent each task using an observation mask $\boldsymbol{\omega}\in\{0,1\}^{3}$, where $\omega_m=1$ indicates that stream $m$ is observed. Text-to-HOI uses $\boldsymbol{\omega}=(0,0,0)$, whereas each completion task observes two streams and generates the remaining one. For an observed stream, we directly use its clean latent with flow time $\tau_m=1$ and set its target velocity to zero. The missing stream follows the same flow used for generation. Let $\mathcal{G}(\boldsymbol{\omega})=\{m\mid\omega_m=0\}$ denote the generated streams. Its flow loss is normalized as
\begin{equation}
\mathcal{L}_{\mathrm{fm}}^{\boldsymbol{\omega}}
=
\frac{3}{|\mathcal{G}(\boldsymbol{\omega})|}
\sum_{m\in\mathcal{G}(\boldsymbol{\omega})}
\mathcal{L}_{m},
\end{equation}
which retains the three-stream loss scale for both generation and completion.

The four tasks are sampled with a ratio of $2{:}1{:}1{:}1$ for Text-to-HOI, body completion, object completion, and hand completion. Observation and task embeddings indicate the available streams and the stream to be generated. A zero-initialized completion adapter is activated only for the missing stream in each transformer block. Because joint attention also updates observed hidden states, we restore them after every block and clamp their latents throughout Euler sampling. The original observed streams are copied into the final motion after decoding, ensuring that only the missing component is modified.

\paragraph{Training Configuration.}
\phantomsection
\label{sec:training_config}
We optimize all models using AdamW with weight decay $10^{-4}$ and gradient clipping at $1.0$. The split VAE is trained for $1{,}000$ epochs with batch size $64$ and learning rate $2\times10^{-4}$. The Text-to-HOI flow is trained for $1000$ epochs with batch size $128$ and learning rate $10^{-4}$. Its learning rate is warmed up over the first $2\%$ of training and then reduced by cosine decay. The joint generation-completion model is also trained  for $1000$ epochs. We set the routing-balance weight to $0.01$ and use a completion-adapter dimension of $128$. Decoded interaction constraints are evaluated on at most $32$ samples using contact-aware windows of $64$ frames and flow times within $[0.2,0.8]$; their weights are linearly warmed up over the first $10\%$ of training. All models are trained with bf16 precision on NVIDIA A800 GPUs. At inference, we integrate the coupled flow using $32$ Euler steps.

\subsection{HOI Understanding and Language Adaptation}
\label{sec:language_supp}

We detail how frozen coupled-flow states are converted into continuous memory tokens and used to adapt the language model for HOI understanding.

\paragraph{HOI Feature Extraction.}
\phantomsection
\label{par:hoi_feature_extraction}
Given a complete interaction $\boldsymbol{x}$, the frozen VAE encodes its body, object, and hand streams into normalized latent vectors $\{\widetilde{\boldsymbol{z}}_m\}_{m\in\{b,o,h\}}$. We pass these latents through the frozen coupled flow transformer at $\tau=1$. Both the text and geometry conditions are replaced with zeros, ensuring that the extracted representation depends only on the interaction motion. Denote the frozen coupled flow transformer before its velocity head by $\mathcal{T}_{\theta}$. The extracted states are:
\begin{equation}
\left(
\boldsymbol{h}_{b}^{L},
\boldsymbol{h}_{o}^{L},
\boldsymbol{h}_{h}^{L}
\right)
=
\mathcal{T}_{\theta}
\left(
\widetilde{\boldsymbol{z}}_{b},
\widetilde{\boldsymbol{z}}_{o},
\widetilde{\boldsymbol{z}}_{h};
1,\boldsymbol{0},\boldsymbol{0}
\right).
\end{equation}
Since each state has exchanged information with the other streams through joint attention, hidden states encode both the motion of stream $m$ and its relation to the complete interaction. We use $\boldsymbol{h}=[\boldsymbol{h}_{b}^{L},\boldsymbol{h}_{o}^{L},\boldsymbol{h}_{h}^{L}]$ as the motion representation for language adaptation.

\paragraph{HOI Language Bridge.}
\phantomsection
\label{par:hoi_language_bridge}
The flow representation contains a number of tokens proportional to the motion duration, whereas the language model requires a compact input with a fixed size. We first add a learned readout stream embedding $\boldsymbol{e}_{m}^{\mathrm{r}}$ to distinguish body, object, and hand tokens:
\begin{equation}
\boldsymbol{h}_{\mathrm{r}}
=
\left[
\boldsymbol{h}_{b}^{L}+\boldsymbol{e}_{b}^{\mathrm{r}};
\boldsymbol{h}_{o}^{L}+\boldsymbol{e}_{o}^{\mathrm{r}};
\boldsymbol{h}_{h}^{L}+\boldsymbol{e}_{h}^{\mathrm{r}}
\right].
\end{equation}
A temporal resampler uses learned slots to attend to the valid tokens in $\boldsymbol{h}_{\mathrm{r}}$. Each resampler block applies cross-attention from the slots to the HOI tokens, followed by slot self-attention and a feed-forward network. We use two blocks, eight attention heads, and $32$ learned slots. The resampled features are mapped into the embedding space of the language model using LayerNorm and a linear projection:
\begin{equation}
\boldsymbol{S}
=
\mathcal{R}
\left(
\boldsymbol{h}_{\mathrm{r}},
\boldsymbol{V}
\right),
\qquad
\boldsymbol{E}
=
\mathcal{P}
\left(
\boldsymbol{S}
\right).
\end{equation}
Here, $\boldsymbol{V}$ masks padded motion tokens, $\boldsymbol{S}$ denotes the resampled slots, and $\boldsymbol{E}$ is the continuous prefix passed to the language model.

\paragraph{Language Model Fine-tuning.}
\phantomsection
\label{par:language_adaptation}
We use the following prompt for all samples:
\begin{quote}
\ttfamily
Object category: [CATEGORY]\\
Question: Describe the human-object interaction.\\
Answer:
\end{quote}
The textual query contains the object category and the instruction \textit{Describe the human-object interaction}. We freeze the VAE, coupled flow transformer, and LLaDA backbone, and optimize the HOI language bridge together with LoRA parameters \citep{lora}. Training uses a masked diffusion objective computed only over answer tokens. For each caption, we mask either a random nonempty subset of its tokens or the complete answer with probability $0.8$, and train the language model to recover the original description.

Raw-motion and flow representations use identical category prompts and contrast the correct HOI memory with a same-category mismatch. Let $\ell_i^{+}$ denote the caption loss using the correct memory and $\ell_i^{-}$ the loss using another interaction from the same category. The complete language objective is:
\begin{equation}
\mathcal{L}_{\mathrm{lang}}
=
\mathcal{L}_{\mathrm{cap}}
+
\lambda_{\mathrm{g}}
\frac{1}{|\mathcal{I}|}
\sum_{i\in\mathcal{I}}
\max
\left(
0,\delta+\ell_i^{+}-\ell_i^{-}
\right),
\end{equation}
where $\mathcal{I}$ contains samples with a valid paired interaction. We set $\lambda_{\mathrm{g}}=0.5$ and $\delta=0.1$. The bridge is first trained alone for $5$ epochs, after which the LoRA parameters are enabled. We train for $100$ epochs with learning rates of $5\times10^{-5}$ for the bridge and $2\times10^{-6}$ for LoRA. LoRA uses rank $8$, scale $16$, and dropout $0.05$. During inference, LLaDA converts the continuous HOI memory into a textual description through $48$ denoising steps.

\section{Additional Experimental Results}
\label{sec:add_exp_supp}

This section reports additional generation and understanding results on OMOMO and BEHAVE, evaluates all three stream completion tasks across datasets, and presents qualitative results for HOI understanding.

\noindent\textbf{HOI Generation on Additional Datasets.}
\begin{table*}
\caption{\footnotesize
\textbf{Quantitative comparisons on the OMOMO dataset between our method and baseline approaches.}
We report R-Precision with a batch size of 256.}
\label{tab:omomo_main}
\centering
\resizebox{\textwidth}{!}{%
\begin{tabular}{@{}lcccccccccccc@{}}
\toprule
\multirow[c]{2}{*}{Method}
& \multicolumn{3}{c}{R-Precision $\uparrow$}
& \multirow[c]{2}{*}{FID $\downarrow$}
& \multirow[c]{2}{*}{MM Dist. $\downarrow$}
& \multirow[c]{2}{*}{Diversity $\rightarrow$}
& \multirow[c]{2}{*}{FSR $\downarrow$}
& \multirow[c]{2}{*}{Pene. $\downarrow$}
& \multirow[c]{2}{*}{Contact $\rightarrow$}
& \multicolumn{3}{c}{Interaction $\uparrow$} \\
\cmidrule(lr){2-4}
\cmidrule(lr){11-13}
& Top 1 & Top 2 & Top 3
& & & & & & &
$C_{\mathrm{prec}}$
& $C_{\mathrm{rec}}$
& $C_{\mathrm{F1}}$ \\
\midrule

Ground Truth
& \omval{0.318}{0.003}
& \omval{0.560}{0.001}
& \omval{0.731}{0.001}
& \omval{0.000}{0.000}
& \omval{1.499}{0.000}
& \omval{7.400}{0.047}
& \omval{0.012}{0.039}
& \omval{0.067}{0.000}
& \omval{0.262}{0.000}
& \omval{1.000}{0.000}
& \omval{1.000}{0.000}
& \omval{1.000}{0.000} \\

\midrule

HOI-Diff
& \omval{0.221}{0.005}
& \omval{0.403}{0.001}
& \omval{0.542}{0.001}
& \omval{0.987}{0.023}
& \omval{2.520}{0.022}
& \omval{7.248}{0.066}
& \omval{0.021}{0.058}
& \omval{0.131}{0.005}
& \omval{0.135}{0.002}
& \omval{0.862}{0.000}
& \omval{0.696}{0.000}
& \omval{0.739}{0.001} \\

CHOIS
& \omval{0.291}{0.000}
& \omval{0.515}{0.009}
& \omval{0.676}{0.008}
& \omval{0.132}{0.016}
& \omval{1.746}{0.002}
& \omval{7.417}{0.104}
& \omval{0.019}{0.053}
& \omval{0.096}{0.004}
& \omsecond{0.230}{0.001}
& \omsecond{0.937}{0.000}
& \omsecond{0.900}{0.004}
& \ombest{0.910}{0.002} \\

InterDiff
& \omval{0.312}{0.003}
& \omval{0.548}{0.004}
& \omval{0.718}{0.004}
& \omval{0.163}{0.009}
& \omval{1.587}{0.006}
& \omval{7.387}{0.045}
& \omval{0.025}{0.062}
& \omval{0.100}{0.002}
& \omval{0.206}{0.000}
& \ombest{0.940}{0.016}
& \omval{0.863}{0.016}
& \omval{0.890}{0.017} \\

Text2HOI
& \omval{0.286}{0.001}
& \omval{0.513}{0.001}
& \omval{0.655}{0.001}
& \omval{0.168}{0.000}
& \omval{1.832}{0.006}
& \omval{7.359}{0.034}
& \omval{0.020}{0.056}
& \omval{0.127}{0.002}
& \omval{0.163}{0.000}
& \omval{0.919}{0.001}
& \omval{0.757}{0.002}
& \omval{0.807}{0.002} \\

ROG
& \omval{0.309}{0.006}
& \omval{0.545}{0.007}
& \omval{0.713}{0.006}
& \omval{0.136}{0.018}
& \omval{1.674}{0.012}
& \ombest{7.408}{0.080}
& \omsecond{0.016}{0.050}
& \omval{0.093}{0.003}
& \omval{0.221}{0.002}
& \omval{0.928}{0.004}
& \omval{0.846}{0.008}
& \omval{0.876}{0.006} \\

LIGHT
& \omval{0.302}{0.004}
& \omval{0.538}{0.015}
& \omval{0.707}{0.011}
& \omval{0.099}{0.018}
& \omval{1.657}{0.010}
& \omval{7.443}{0.043}
& \ombest{0.013}{0.045}
& \omval{0.100}{0.001}
& \omval{0.194}{0.000}
& \ombest{0.940}{0.000}
& \omval{0.851}{0.008}
& \omval{0.881}{0.005} \\

\midrule

\text{TRACE}
& \omsecond{0.342}{0.003}
& \omsecond{0.601}{0.006}
& \omsecond{0.764}{0.002}
& \omsecond{0.087}{0.009}
& \omsecond{1.557}{0.017}
& \omval{7.422}{0.122}
& \omval{0.018}{0.054}
& \omsecond{0.089}{0.005}
& \omval{0.227}{0.002}
& \omval{0.917}{0.009}
& \omval{0.889}{0.009}
& \omval{0.894}{0.010} \\

\text{$\text{TRACE}^{\text{ joint}}$}
& \ombest{0.344}{0.004}
& \ombest{0.611}{0.007}
& \ombest{0.768}{0.004}
& \ombest{0.077}{0.008}
& \ombest{1.541}{0.012}
& \omval{7.433}{0.099}
& \omval{0.018}{0.042}
& \ombest{0.084}{0.002}
& \ombest{0.232}{0.001}
& \omval{0.919}{0.005}
& \ombest{0.901}{0.003}
& \omsecond{0.904}{0.004} \\

\bottomrule
\end{tabular}%
}
\end{table*}
\begin{table*}[t]
\centering
\caption{\footnotesize
\textbf{Quantitative comparisons on the BEHAVE dataset between our method and baseline approaches.} We report R-Precision with a batch size of 256.}
\label{tab:behave_main}
\setlength{\tabcolsep}{3.5pt}
\resizebox{\textwidth}{!}{%
\begin{tabular}{@{}lcccccccccccc@{}}
\toprule
\multirow[c]{2}{*}{Method}
& \multicolumn{3}{c}{R-Precision $\uparrow$}
& \multirow[c]{2}{*}{FID $\downarrow$}
& \multirow[c]{2}{*}{MM Dist. $\downarrow$}
& \multirow[c]{2}{*}{Diversity $\rightarrow$}
& \multirow[c]{2}{*}{FSR $\downarrow$}
& \multirow[c]{2}{*}{Pene. $\downarrow$}
& \multirow[c]{2}{*}{Contact $\rightarrow$}
& \multicolumn{3}{c}{Interaction $\uparrow$} \\
\cmidrule(lr){2-4}
\cmidrule(lr){11-13}
& Top 1 & Top 2 & Top 3
& \multicolumn{6}{c}{}
& $C_{\mathrm{prec}}$
& $C_{\mathrm{rec}}$
& $C_{\mathrm{F1}}$ \\
\midrule

Ground Truth
& \omval{0.559}{0.005}
& \omval{0.910}{0.000}
& \omval{0.926}{0.000}
& \omval{0.000}{0.000}
& \omval{1.483}{0.018}
& \omval{6.940}{0.153}
& \omval{0.181}{0.224}
& \omval{0.094}{0.001}
& \omval{0.204}{0.001}
& \omval{1.000}{0.000}
& \omval{1.000}{0.000}
& \omval{1.000}{0.000} \\

\midrule

HOI-Diff
& \omval{0.167}{0.009}
& \omval{0.298}{0.012}
& \omval{0.358}{0.018}
& \omval{1.463}{0.044}
& \omval{4.152}{0.038}
& \omval{6.723}{0.074}
& \omval{0.179}{0.238}
& \omval{0.171}{0.007}
& \omval{0.100}{0.003}
& \omval{0.646}{0.008}
& \omval{0.515}{0.010}
& \omval{0.519}{0.007} \\

CHOIS
& \omval{0.219}{0.000}
& \omval{0.396}{0.003}
& \omval{0.492}{0.022}
& \omval{0.705}{0.027}
& \omval{3.423}{0.045}
& \omval{6.863}{0.061}
& \omval{0.191}{0.253}
& \omval{0.128}{0.011}
& \omsecond{0.170}{0.003}
& \omval{0.717}{0.018}
& \omval{0.662}{0.000}
& \omval{0.650}{0.005} \\

InterDiff
& \omval{0.258}{0.027}
& \omval{0.443}{0.019}
& \omval{0.531}{0.027}
& \omval{0.630}{0.033}
& \omval{3.219}{0.040}
& \omval{6.862}{0.066}
& \omval{0.208}{0.264}
& \omval{0.143}{0.009}
& \ombest{0.173}{0.005}
& \omval{0.723}{0.001}
& \omsecond{0.682}{0.013}
& \omval{0.676}{0.004} \\

Text2HOI
& \omval{0.135}{0.003}
& \omval{0.242}{0.000}
& \omval{0.322}{0.008}
& \omval{1.291}{0.073}
& \omval{3.815}{0.026}
& \omval{6.544}{0.013}
& \omval{0.177}{0.233}
& \omval{0.179}{0.003}
& \omval{0.128}{0.001}
& \omval{0.703}{0.011}
& \omval{0.579}{0.007}
& \omval{0.587}{0.008} \\

ROG
& \omval{0.351}{0.010}
& \omval{0.528}{0.013}
& \omval{0.620}{0.016}
& \omval{0.612}{0.046}
& \omval{3.169}{0.031}
& \ombest{6.948}{0.084}
& \ombest{0.168}{0.231}
& \ombest{0.106}{0.006}
& \omval{0.151}{0.003}
& \omsecond{0.749}{0.006}
& \omval{0.628}{0.010}
& \omval{0.656}{0.008} \\

LIGHT
& \omval{0.277}{0.011}
& \omval{0.475}{0.014}
& \omval{0.561}{0.041}
& \ombest{0.481}{0.087}
& \omval{3.277}{0.076}
& \omsecond{6.961}{0.066}
& \omval{0.171}{0.245}
& \omval{0.140}{0.004}
& \omsecond{0.170}{0.001}
& \omval{0.731}{0.006}
& \ombest{0.720}{0.011}
& \ombest{0.697}{0.007} \\

\midrule

\text{TRACE}
& \omsecond{0.438}{0.009}
& \omsecond{0.584}{0.004}
& \omsecond{0.681}{0.009}
& \omval{0.688}{0.001}
& \omsecond{3.145}{0.011}
& \omval{6.978}{0.447}
& \omsecond{0.169}{0.001}
& \omval{0.163}{0.006}
& \omval{0.141}{0.003}
& \omval{0.737}{0.002}
& \omval{0.650}{0.002}
& \omval{0.653}{0.003} \\

\text{TRACE$^{\text{ joint}}$}
& \ombest{0.466}{0.004}
& \ombest{0.613}{0.017}
& \ombest{0.688}{0.017}
& \omsecond{0.578}{0.013}
& \ombest{3.044}{0.001}
& \omval{6.906}{0.357}
& \omval{0.175}{0.001}
& \omsecond{0.122}{0.006}
& \omval{0.161}{0.001}
& \ombest{0.758}{0.009}
& \omsecond{0.682}{0.002}
& \omsecond{0.682}{0.006} \\

\bottomrule
\end{tabular}%
}
\end{table*}
Tables \ref{tab:omomo_main} and \ref{tab:behave_main} report generation results on OMOMO and BEHAVE, respectively. Following prior methods, all models are trained separately on each dataset using the corresponding training split. TRACE performs favorably on most generation metrics, producing lower feature distribution distances and stronger alignment between generated motion and text. The improvements in contact and physical metrics further indicate that the generated human and object trajectories remain coordinated. Although OMOMO and BEHAVE contain substantially fewer interactions than InterAct, the coupled flow can still learn effective dependencies among body, object, and hand motion. These results show that TRACE remains effective under different dataset scales and interaction distributions.

\begin{table*}[t]
    \caption{\footnotesize \textbf{Comparison of HOI understanding on OMOMO.}
    Our model is trained for 100 epochs.}
    \label{tab:omomo_understanding}
    \centering
    \resizebox{\textwidth}{!}{%
    \begin{tabular}{l c c c c c c c c c}
    \toprule
    \multirow[c]{2}{*}{Methods}
    & \multicolumn{3}{c}{R-Precision $\uparrow$}
    & \multirow[c]{2}{*}{MM Dist. $\downarrow$}
    & \multirow[c]{2}{*}{Bleu@1 $\uparrow$}
    & \multirow[c]{2}{*}{Bleu@4 $\uparrow$}
    & \multirow[c]{2}{*}{Rouge $\uparrow$}
    & \multirow[c]{2}{*}{Cider $\uparrow$}
    & \multirow[c]{2}{*}{BertScore $\uparrow$} \\
    \cmidrule(lr){2-4}
    & Top 1 & Top 2 & Top 3
    & & & & & & \\
    \midrule

    Real
    & 0.345 & 0.609 & 0.778
    & 1.499
    & --
    & --
    & --
    & --
    & -- \\

    \midrule

    TM2T
    & 0.231
    & 0.421
    & 0.598
    & 2.173
    & 80.62
    & 74.18
    & 90.87
    & 7.46
    & 91.72 \\

    MotionGPT
    & 0.254
    & 0.459
    & 0.632
    & 2.046
    & 84.51
    & 79.36
    & 91.92
    & 8.16
    & 92.31 \\

    LaMPM2T
    & 0.276
    & 0.490
    & 0.661
    & 1.928
    & 88.74
    & \underline{85.61}
    & 92.96
    & \textbf{9.37}
    & \underline{93.21} \\

    HOIGPT
    & \underline{0.291}
    & \underline{0.522}
    & \underline{0.688}
    & \underline{1.824}
    & \underline{91.56}
    & 85.18
    & \underline{93.58}
    & \underline{9.08}
    & 93.14 \\

    \midrule

    \text{\text{TRACE}$^\text{ Raw}$}
    & 0.260 
    & 0.472 
    & 0.658 
    & 1.965
    & 86.29
    & 82.91 
    & 92.48 
    & 7.31 
    & 93.18 \\

    \text{TRACE}
    & \textbf{0.316}
    & \textbf{0.553}
    & \textbf{0.715}
    & \textbf{1.692}
    & \textbf{93.77}
    & \textbf{90.12}
    & \textbf{94.20}
    & 8.80
    & \textbf{93.30} \\

    \bottomrule
    \end{tabular}%
    }
    \vspace{-2mm}
\end{table*}
\begin{table*}
    \caption{\footnotesize \textbf{Comparison of HOI understanding on BEHAVE.}
    Our model is trained for 100 epochs.}
    \label{tab:behave_understanding}
    \centering
    \resizebox{\textwidth}{!}{%
    \begin{tabular}{l c c c c c c c c c}
    \toprule
    \multirow[c]{2}{*}{Methods}
    & \multicolumn{3}{c}{R-Precision $\uparrow$}
    & \multirow[c]{2}{*}{MM Dist. $\downarrow$}
    & \multirow[c]{2}{*}{Bleu@1 $\uparrow$}
    & \multirow[c]{2}{*}{Bleu@4 $\uparrow$}
    & \multirow[c]{2}{*}{Rouge $\uparrow$}
    & \multirow[c]{2}{*}{Cider $\uparrow$}
    & \multirow[c]{2}{*}{BertScore $\uparrow$} \\
    \cmidrule(lr){2-4}
    & Top 1 & Top 2 & Top 3
    & & & & & & \\
    \midrule

    Real
    & 0.913 & 0.944 & 0.956
    & 1.498
    & --
    & --
    & --
    & --
    & -- \\

    \midrule

    TM2T
    & 0.246 & 0.383 & 0.501
    & 4.143
    & 23.18
    & 9.62
    & 37.16
    & 4.9
    & 22.18 \\

    MotionGPT
    & 0.276 & 0.421 & 0.544
    & 3.846
    & 26.43
    & 11.57
    & 40.13
    & 5.9
    & 24.16 \\

    LaMPM2T
    & 0.306
    & 0.459
    & 0.586
    & 3.596
    & 29.14
    & \underline{13.92}
    & 43.09
    & 6.7
    & \underline{26.12} \\

    HOIGPT
    & \underline{0.336}
    & \underline{0.497}
    & \textbf{0.632}
    & \textbf{3.301}
    & \textbf{32.64}
    & \textbf{16.52}
    & \textbf{46.57}
    & \textbf{8.32}
    & \textbf{30.61} \\

    \midrule

    \text{\text{TRACE}$^\text{ Raw}$}
    & 0.290 
    & 0.438 
    & 0.555 
    & 3.756
    & 27.52 
    & 12.27 
    & 41.29 
    & 6.1 
    & 24.35 \\

    \text{TRACE}
    & \textbf{0.350}
    & \textbf{0.513}
    & \underline{0.619}
    & \underline{3.484}
    & \underline{30.58}
    & 13.83
    & \underline{44.34}
    & \underline{7.4}
    & 25.95 \\

    \bottomrule
    \end{tabular}%
    }
    \vspace{-2mm}
\end{table*}

\noindent\textbf{HOI Understanding on Additional Datasets.}
Tables \ref{tab:omomo_understanding} and~\ref{tab:behave_understanding} compare understanding on OMOMO and BEHAVE. Across both datasets, frozen coupled-flow features improve every reported metric over raw-motion encoding. Against external baselines, performance varies by dataset: TRACE is competitive on OMOMO and improves R@1 and R@2 on BEHAVE, but does not lead most BEHAVE language metrics. These results support the internal representation comparison without implying uniform superiority over all baselines.

\begin{table*}[h!]
\centering
\caption{\textbf{Stream completion on InterAct, OMOMO, and BEHAVE.} For each dataset, one jointly trained checkpoint is evaluated under three conditioning settings.}
\label{tab:any_stream}
\small
\setlength{\tabcolsep}{4pt}
\renewcommand{\arraystretch}{0.80}
\resizebox{\textwidth}{!}{%
\begin{tabular}{lllcccccccc}
\toprule
\multirow{2}{*}{Dataset}
& \multirow{2}{*}{Setting}
& \multirow{2}{*}{Target}
& \multicolumn{3}{c}{Human reconstruction $\downarrow$}
& \multicolumn{2}{c}{Object reconstruction $\downarrow$}
& \multicolumn{3}{c}{Interaction $\uparrow$} \\
\cmidrule(lr){4-6}
\cmidrule(lr){7-8}
\cmidrule(lr){9-11}
& & &
B-MPJPE & Root & H-JPE &
Center & V2V &
$C_{\rm prec}$ & $C_{\rm rec}$ & $C_{\rm F1}$ \\
\midrule


\multirow{9}{*}{InterAct}
& \multirow{3}{*}{Text/BPS only}
& Object & -- & -- & -- & 69.90 & 74.79 & 0.420 & 0.172 & 0.195 \\
& & Hands & -- & -- & 726.46 & -- & -- & 0.390 & 0.148 & 0.164 \\
& & Body & 170.74 & 574.55 & -- & -- & -- & -- & -- & -- \\
\cmidrule(lr){2-11}

& \multirow{3}{*}{Streams only}
& Object
& -- & -- & --
& \underline{19.25}
& \underline{34.77}
& \underline{0.503}
& \underline{0.309}
& \underline{0.330} \\
& & Hands
& -- & --
& \textbf{44.91}
& -- & --
& \underline{0.669}
& \underline{0.613}
& \underline{0.610} \\
& & Body
& \textbf{55.21}
& \textbf{72.89}
& -- & -- & -- & -- & -- & -- \\
\cmidrule(lr){2-11}

& \multirow{3}{*}{Full}
& Object
& -- & -- & --
& \textbf{15.34}
& \textbf{26.51}
& \textbf{0.584}
& \textbf{0.411}
& \textbf{0.434} \\
& & Hands
& -- & --
& \underline{47.19}
& -- & --
& \textbf{0.679}
& \textbf{0.629}
& \textbf{0.625} \\
& & Body
& \underline{60.74}
& \underline{80.01}
& -- & -- & -- & -- & -- & -- \\

\midrule


\multirow{9}{*}{OMOMO}
& \multirow{3}{*}{Text/BPS only}
& Object & -- & -- & -- & 95.44 & 100.47 & 0.407 & 0.093 & 0.128 \\
& & Hands & -- & -- & 969.38 & -- & -- & 0.379 & 0.075 & 0.107 \\
& & Body & 200.85 & 781.40 & -- & -- & -- & -- & -- & -- \\
\cmidrule(lr){2-11}

& \multirow{3}{*}{Streams only}
& Object
& -- & -- & --
& \underline{18.25}
& \underline{36.37}
& \underline{0.530}
& \underline{0.324}
& \underline{0.354} \\
& & Hands
& -- & --
& \textbf{69.75}
& -- & --
& \textbf{0.669}
& \textbf{0.595}
& \textbf{0.594} \\
& & Body
& \textbf{50.89}
& \textbf{83.30}
& -- & -- & -- & -- & -- & -- \\
\cmidrule(lr){2-11}

& \multirow{3}{*}{Full}
& Object
& -- & -- & --
& \textbf{14.16}
& \textbf{25.30}
& \textbf{0.612}
& \textbf{0.414}
& \textbf{0.445} \\
& & Hands
& -- & --
& \underline{71.34}
& -- & --
& \underline{0.665}
& \underline{0.592}
& \underline{0.592} \\
& & Body
& \underline{51.80}
& \underline{83.81}
& -- & -- & -- & -- & -- & -- \\

\midrule


\multirow{9}{*}{BEHAVE}
& \multirow{3}{*}{Text/BPS only}
& Object & -- & -- & -- & 52.70 & 60.37 & 0.382 & 0.135 & 0.155 \\
& & Hands & -- & -- & 580.85 & -- & -- & 0.354 & 0.113 & 0.129 \\
& & Body & 183.56 & 419.61 & -- & -- & -- & -- & -- & -- \\
\cmidrule(lr){2-11}

& \multirow{3}{*}{Streams only}
& Object
& -- & -- & --
& \underline{21.99}
& \underline{37.55}
& \underline{0.407}
& \underline{0.261}
& \underline{0.275} \\
& & Hands
& -- & --
& \textbf{61.76}
& -- & --
& \underline{0.586}
& \textbf{0.540}
& \textbf{0.543} \\
& & Body
& \textbf{68.75}
& \textbf{80.87}
& -- & -- & -- & -- & -- & -- \\
\cmidrule(lr){2-11}

& \multirow{3}{*}{Full}
& Object
& -- & -- & --
& \textbf{18.58}
& \textbf{31.34}
& \textbf{0.484}
& \textbf{0.322}
& \textbf{0.342} \\
& & Hands
& -- & --
& \underline{76.61}
& -- & --
& \textbf{0.589}
& \underline{0.533}
& \underline{0.535} \\
& & Body
& \underline{68.87}
& \underline{81.30}
& -- & -- & -- & -- & -- & -- \\

\bottomrule
\end{tabular}%
}
\end{table*}

\noindent\textbf{Stream Completion on Three Datasets.}
Table~\ref{tab:any_stream} compares three conditioning settings for object, hand, and body completion. Text/BPS-only uses no observed motion, Streams-only removes the text and BPS embeddings, and Full uses all conditions. For Text/BPS-only, the generated target is evaluated with the synchronized ground-truth non-target streams. In the other settings, observed streams are copied unchanged. Each dataset uses one TRACE checkpoint, with results averaged over two evaluation seeds.

\textbf{B-MPJPE} measures root-relative body-joint error, \textbf{Root} measures root-translation error, and \textbf{H-JPE} measures hand-joint error, all in mm. \textbf{Center} and \textbf{V2V} measure object-center and surface errors in cm. $C_{\rm prec}$, $C_{\rm rec}$, and $C_{\rm F1}$ are macro-averaged either-hand contact metrics under a 5cm threshold; contact is omitted when both hand and object are observed.

Across all datasets, observed streams provide the main completion signal. Full conditioning consistently yields the lowest object reconstruction errors and the highest contact scores for object completion. For hand and body completion, Streams-only remains comparable and sometimes produces lower paired reconstruction errors, showing that the observed trajectories already provide most of the required motion information.

\begin{figure*}
    \centering
    \includegraphics[width=\textwidth]{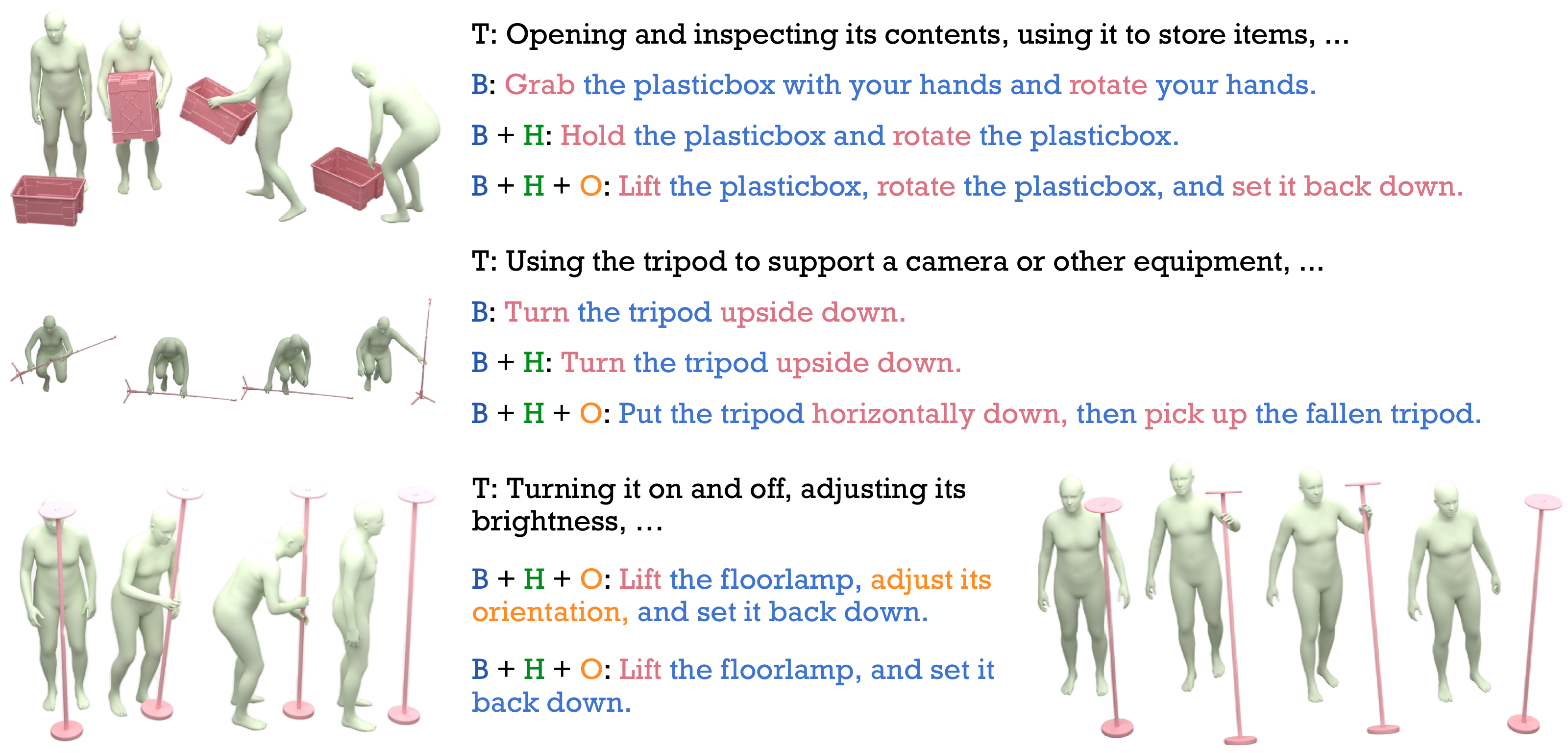}
    \caption{\footnotesize
    \textbf{Qualitative HOI understanding.}
    Adding hand and object evidence produces more complete interaction descriptions. $T$ denotes the reference description.}
    \label{fig:understanding}
\end{figure*}

\noindent\textbf{Qualitative Evaluation of HOI Understanding.}
Figure~\ref{fig:understanding} analyzes the effects of motion evidence and representation on HOI understanding. In the plastic box and tripod examples, partial body and hand observations yield coarse descriptions, while including all three streams recovers more complete object movement and temporal progression. The floor lamp example further compares two representations under the same complete input. Coupled flow features recognize the orientation adjustment, whereas direct raw motion encoding omits this detail. These results show that both complete HOI evidence and the interaction relations learned during generation benefit HOI understanding.

\begin{table*}[h!]
\caption{\footnotesize
\textbf{Ablation study of stream modeling on OMOMO.}
We report R-Precision with a batch size of 256.}
\label{tab:ablation_stream_omomo}
\centering
\scriptsize
\setlength{\tabcolsep}{2.6pt}
\renewcommand{\arraystretch}{1.05}
\resizebox{\textwidth}{!}{%
\begin{tabular}{@{}l*{12}{c}@{}}
\toprule
\multirow[c]{2}{*}{Strategy}
& \multicolumn{3}{c}{R-Precision $\uparrow$}
& \multirow[c]{2}{*}{FID $\downarrow$}
& \multirow[c]{2}{*}{MM Dist. $\downarrow$}
& \multirow[c]{2}{*}{Diversity $\rightarrow$}
& \multirow[c]{2}{*}{FSR $\downarrow$}
& \multirow[c]{2}{*}{Pene. $\downarrow$}
& \multirow[c]{2}{*}{Contact $\rightarrow$}
& \multicolumn{3}{c}{Interaction $\uparrow$} \\
\cmidrule(lr){2-4}
\cmidrule(lr){11-13}
& Top 1 & Top 2 & Top 3
& & & & & & &
$C_{\mathrm{prec}}$
& $C_{\mathrm{rec}}$
& $C_{F1}$ \\
\midrule

GT
& \omval{0.318}{0.003}
& \omval{0.560}{0.001}
& \omval{0.731}{0.001}
& \omval{0.000}{0.000}
& \omval{1.499}{0.000}
& \omval{7.400}{0.047}
& \omval{0.012}{0.039}
& \omval{0.067}{0.000}
& \omval{0.262}{0.000}
& \omval{1.000}{0.000}
& \omval{1.000}{0.000}
& \omval{1.000}{0.000} \\

\midrule

Independent
& \omval{0.325}{0.004}
& \omval{0.570}{0.007}
& \omval{0.731}{0.004}
& \omval{0.120}{0.011}
& \omval{1.701}{0.020}
& \omval{7.467}{0.124}
& \ombest{0.015}{0.078}
& \ombest{0.077}{0.006}
& \omval{0.174}{0.002}
& \omval{0.849}{0.011}
& \omval{0.651}{0.012}
& \omval{0.696}{0.012} \\

Unified
& \omsecond{0.339}{0.003}
& \omsecond{0.597}{0.006}
& \omsecond{0.759}{0.003}
& \omsecond{0.093}{0.010}
& \omsecond{1.590}{0.018}
& \ombest{7.411}{0.118}
& \omval{0.022}{0.032}
& \omval{0.097}{0.006}
& \omsecond{0.218}{0.002}
& \omsecond{0.906}{0.010}
& \omsecond{0.842}{0.011}
& \omsecond{0.855}{0.011} \\

Coupled
& \ombest{0.342}{0.003}
& \ombest{0.601}{0.006}
& \ombest{0.764}{0.002}
& \ombest{0.087}{0.009}
& \ombest{1.557}{0.017}
& \omsecond{7.422}{0.122}
& \omsecond{0.018}{0.054}
& \omsecond{0.089}{0.005}
& \ombest{0.227}{0.002}
& \ombest{0.917}{0.009}
& \ombest{0.889}{0.009}
& \ombest{0.894}{0.010} \\

\bottomrule
\end{tabular}%
}
\end{table*}
\begin{table*}[h!]
\caption{\footnotesize
\textbf{Ablation study of stream modeling on BEHAVE.}
We report R-Precision with a batch size of 256.}
\label{tab:ablation_stream_behave}
\centering
\scriptsize
\setlength{\tabcolsep}{2.6pt}
\renewcommand{\arraystretch}{1.05}
\resizebox{\textwidth}{!}{%
        \begin{tabular}{@{}l*{12}{c}@{}}
        \toprule
        \multirow[c]{2}{*}{Strategy}
        & \multicolumn{3}{c}{R-Precision $\uparrow$}
        & \multirow[c]{2}{*}{FID $\downarrow$}
        & \multirow[c]{2}{*}{MM Dist. $\downarrow$}
        & \multirow[c]{2}{*}{Diversity $\rightarrow$}
        & \multirow[c]{2}{*}{FSR $\downarrow$}
        & \multirow[c]{2}{*}{Pene. $\downarrow$}
        & \multirow[c]{2}{*}{Contact $\rightarrow$}
        & \multicolumn{3}{c}{Interaction $\uparrow$} \\
        \cmidrule(lr){2-4}
        \cmidrule(lr){11-13}
        & Top 1 & Top 2 & Top 3
        & & & & & & &
        $C_{\mathrm{prec}}$
        & $C_{\mathrm{rec}}$
        & $C_{F1}$ \\
        \midrule
        
        GT
        & \omval{0.559}{0.005}
        & \omval{0.910}{0.000}
        & \omval{0.926}{0.000}
        & \omval{0.000}{0.000}
        & \omval{1.483}{0.018}
        & \omval{6.940}{0.153}
        & \omval{0.181}{0.224}
        & \omval{0.094}{0.001}
        & \omval{0.204}{0.001}
        & \omval{1.000}{0.000}
        & \omval{1.000}{0.000}
        & \omval{1.000}{0.000} \\
        
        \midrule

        Independent
        & \omval{0.414}{0.011}
        & \omval{0.550}{0.007}
        & \omval{0.643}{0.011}
        & \omval{0.842}{0.014}
        & \omval{3.392}{0.048}
        & \omval{7.056}{0.051}
        & \omval{0.194}{0.122}
        & \ombest{0.142}{0.007}
        & \omval{0.104}{0.003}
        & \omval{0.676}{0.004}
        & \omval{0.451}{0.005}
        & \omval{0.486}{0.005} \\
        
        Unified
        & \omsecond{0.435}{0.010}
        & \omsecond{0.580}{0.006}
        & \omsecond{0.678}{0.010}
        & \omsecond{0.716}{0.012}
        & \omsecond{3.188}{0.046}
        & \ombest{6.956}{0.030}
        & \omsecond{0.177}{0.194}
        & \omval{0.174}{0.007}
        & \omsecond{0.136}{0.003}
        & \omsecond{0.730}{0.003}
        & \omsecond{0.622}{0.004}
        & \omsecond{0.631}{0.004} \\
        
        Coupled
        & \ombest{0.438}{0.009}
        & \ombest{0.584}{0.004}
        & \ombest{0.681}{0.009}
        & \ombest{0.688}{0.017}
        & \ombest{3.145}{0.040}
        & \omsecond{6.978}{0.047}
        & \ombest{0.169}{0.139}
        & \omsecond{0.163}{0.006}
        & \ombest{0.141}{0.003}
        & \ombest{0.737}{0.002}
        & \ombest{0.650}{0.009}
        & \ombest{0.653}{0.003} \\
        
        \bottomrule
        \end{tabular}%
        }
\end{table*}

\noindent\textbf{Impact of Stream Modeling on Additional Datasets.}
Tables \ref{tab:ablation_stream_omomo} and \ref{tab:ablation_stream_behave} extend the comparison of unified, independent, and coupled stream modeling to OMOMO and BEHAVE. The coupled formulation achieves the strongest overall generation and interaction results on both datasets, consistent with the InterAct results in the main paper. This confirms that preserving separate stream states while coupling their evolution remains effective across different dataset scales and interaction distributions.

\begin{table}[h!]
\centering
\caption{\textbf{Inference efficiency on InterAct.}
All methods generate a batch of 64 interactions, each containing 300 frames, on one NVIDIA A800 GPU using BF16.}
\label{tab:runtime}
\small
\setlength{\tabcolsep}{5pt}
\begin{tabular}{lccc}
\toprule
Method
& Time / batch (s) $\downarrow$
& Time / sequence (s) $\downarrow$
& Relative latency ($\times$) $\downarrow$ \\
\midrule
HOI-Diff
& $35.20^{\pm 0.12}$
& $0.550$
& $10.2$ \\

InterDiff
& $42.88^{\pm 0.35}$
& $0.670$
& $12.5$ \\

LIGHT
& $192.76^{\pm 0.57}$
& $3.012$
& $56.0$ \\

TRACE
& $3.44^{\pm 0.01}$
& $0.054$
& $1.0$ \\
\bottomrule
\end{tabular}
\end{table}

\paragraph{Inference efficiency.}
We evaluate all methods on one NVIDIA A800 GPU using BF16, a batch size of $64$, and sequences of $300$ frames. Each method uses its official inference configuration. After five warm-up runs, we measure $10$ runs with CUDA synchronization before and after timing. Runtime includes sampling and motion decoding but excludes data loading, rendering, and file I/O.
As shown in Table \ref{tab:runtime}, TRACE requires 3.44 seconds per batch and is $10.2\times$, $12.5\times$, and $56.0\times$ faster than HOI-Diff, InterDiff, and LIGHT, respectively.


\end{document}